\PassOptionsToPackage{table}{xcolor}
\documentclass{article}
\usepackage{iclr2027_conference,times}

\usepackage[utf8]{inputenc}
\usepackage[T1]{fontenc}
\usepackage{hyperref}
\usepackage{url}
\usepackage{booktabs}
\usepackage{amsfonts}
\usepackage{nicefrac}
\usepackage{microtype}
\usepackage{graphicx}
\usepackage{subcaption}
\usepackage{multirow}
\usepackage{adjustbox}
\usepackage{float}
\usepackage{placeins}
\usepackage{amsmath}
\usepackage{amssymb}
\usepackage{amsthm}
\usepackage{algorithm}
\usepackage{algorithmic}
\usepackage{colortbl}
\usepackage{wrapfig}

\title{Balancing Representations by Identifying and Generating Under-represented Data}
\author{Filipe Laitenberger\textsuperscript{1},
Hirokatsu Kataoka\textsuperscript{2},
Yuki M. Asano\textsuperscript{3},\\
\bf
Cees G. M. Snoek\textsuperscript{4},
Mohammadreza Salehi\textsuperscript{4}\\
\small
\textsuperscript{1}Humboldt-Universit\"at zu Berlin \ \ \
\textsuperscript{2}University of Oxford \ \ \
\textsuperscript{3}University of Technology Nuremberg\\
\small
\textsuperscript{4}University of Amsterdam\\
\small
Correspondence to: \texttt{filipe.laitenberger@hu-berlin.de}
}

\iclrpreprinttrue

\newcommand{\method}{BRIDGE}
\newtheorem{proposition}{Proposition}

\begin{document}

\maketitle

\begin{abstract}
Self-supervised learning inherits the coverage of its source data. When an unlabeled corpus is long-tailed, the representation becomes dense around frequent modes and thin where coverage is weak. Objectives for long-tailed SSL reweight the images already in the corpus and add no new ones. We introduce \method\ (\textbf{B}alancing \textbf{R}epresentations by \textbf{Id}entifying and \textbf{Ge}nerating Under-represented Data), a label-free feedback loop around the source set. The current encoder scores every source image by its mean $k$NN distance, and a budgeted acquisition rule selects images from a band of that score distribution where support is thin and the generator still produces faithful variants. A frozen image-to-image diffusion model then adds local variants of the selected anchors, and pretraining continues on the enlarged source set. We formulate acquisition as recoverability-weighted allocation of local support and derive when a finite budget should favor a sparse band that excludes the extreme tail and when reallocation after each repair is optimal under a fixed objective. Across five long-tailed, web-scraped and synthetic source regimes, \method\ has the highest seven-task linear-probe and $k$NN average of the compared methods, and paired experiments extend the gain to ViT-S under contrastive and self-distillation objectives, to low-label fine-tuning and to semantic segmentation. The linear-probe gain is smaller under masked reconstruction. Generated repair matches an oracle that restores the withheld real images, and under a frozen encoder the inserted images shrink the neighborhoods of the selected anchors about twice as much as those of matched unselected images. \method\ needs no labels and leaves the pretraining objective unchanged, so it can be added to pretraining on uncurated web data, where the encoder decides which images to generate.
\end{abstract}

\begin{figure}[H]
\centering
\includegraphics[width=0.89\linewidth]{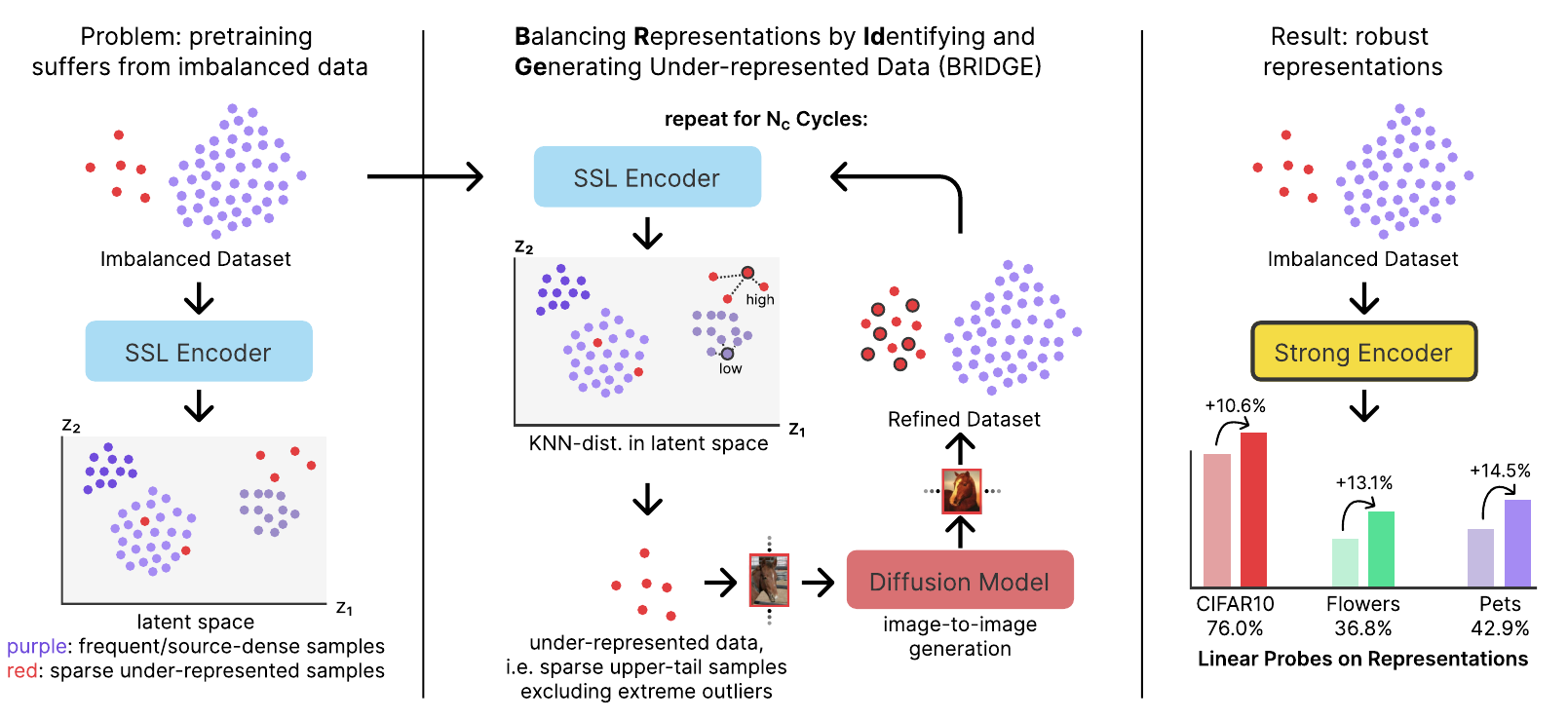}
\caption{Overview of \method. Left, an imbalanced source set embedded by the SSL encoder, with sparse under-represented samples in red. Right, the loop. Distance to the $k$ nearest neighbors in the current latent space selects sparse samples from the upper tail of the score distribution while the extreme tail is excluded, image-to-image generation adds variants of them, and the refined dataset is the source set of the next cycle.}
\label{fig:bridge-overview}
\end{figure}

\section{Introduction}
Self-supervised learning produces transferable visual representations from unlabeled images~\citep{chen2020simple,he2020momentum,DINO,he2022masked,oquab2023dinov2,dinov3}, and every objective learns from what its source corpus contains. When that corpus is long-tailed, the representation concentrates around frequent modes, covers rare regions weakly and transfers worse~\citep{tian2021divide,assran2210hidden,ts,sdclr}. Prior work responds on the model side, with a temperature schedule~\citep{ts}, a pruned self-competitor~\citep{sdclr} or imported external data~\citep{colt}, leaving the source set fixed.

We ask whether the source set can be repaired while learning proceeds. \method, \textbf{B}alancing \textbf{R}epresentations by \textbf{Id}entifying and \textbf{Ge}nerating Under-represented Data, embeds the source set with the current encoder after each cycle and scores every image by its mean distance to its $k$ nearest neighbors. It selects a sparse but semantically coherent band of that distribution, generates local variants of the selected images with a frozen image-to-image diffusion model and continues pretraining on the enlarged set (Figure~\ref{fig:bridge-overview}), while no labels, class frequencies or downstream signals enter the loop.

We formalize acquisition as budgeted allocation over regions of representation space with diminishing returns to local support and repair fidelity that falls in the extreme tail. Under this model the budget belongs in a sparse band that excludes the extreme tail, and greedy reallocation after each repair is optimal under a fixed objective. In the experiments this band beats uniform acquisition by 6.6 to 6.8 points and top-tail acquisition by 7.8 to 8.0 points.

Across five long-tailed, web-scraped and synthetic sources, \method\ raises the seven-task linear-probe average of SimCLR on the same source by 2.4 to 10.6 points and has the best average of all compared methods under linear-probe and $k$NN evaluation. With a power-law class imbalance imposed on ImageNet-100, repair adds 11.1 points, while a class-balanced source of the same size adds 0.1. On ViT-S the linear-probe average rises by 2.0 to 2.8 points under SimCLR, MoCo v3 and DINO, and generated repair reaches the accuracy of an oracle that adds real images withheld from the source, 41.5 against 40.7.

Our contributions are as follows.
\begin{itemize}
\setlength{\itemsep}{1pt}
\item A theory of label-free data repair as recoverability-weighted allocation of local support (Section~\ref{sec:theory}), which shows that under diminishing returns and fidelity that decays with sparsity the budget belongs in a sparse band that excludes the extreme tail, and that greedy reallocation is optimal under a fixed separable objective.
\item \method, a label-free feedback loop that instantiates this policy (Section~\ref{sec:method}) with mean $k$NN distance as the support estimate, a trimmed score band that holds sparse images the generator can still vary faithfully, farthest-point sampling against redundancy and a frozen image-to-image generator as the repair operator.
\item An evaluation on five source regimes and seven transfer tasks under linear-probe and $k$NN protocols (Section~\ref{sec:results}), with paired ViT-S experiments across four SSL objectives, low-label fine-tuning, where accuracy rises by 2.0 and 3.2 points, and PASCAL VOC segmentation, where it rises by 2.4 mIoU.
\item An account of where the effect comes from (Sections~\ref{sec:ablations} and~\ref{sec:geometry}). The band beats uniform and top-tail acquisition as Section~\ref{sec:theory} predicts, and a fixed-encoder test shows the generated images shrinking the neighborhoods of their own anchors twice as much as those of matched unselected images.
\end{itemize}

\section{Related Work}
\label{sec:background}
\paragraph{Self-supervised learning under skewed sources.}
Contrastive, self-distillation and masked-reconstruction pretraining~\citep{chen2020simple,he2020momentum,DINO,oquab2023dinov2,dinov3,he2022masked} is sensitive to source imbalance, which has been addressed with clustered experts, the hidden uniform cluster prior and reweighting by estimated class frequency~\citep{tian2021divide,assran2210hidden,fassl}. The closest methods change the objective on a fixed corpus, SDCLR with a pruned self-competitor~\citep{sdclr} and Temperature Schedules by emphasizing head and tail at different times~\citep{ts}, or import external images~\citep{colt}.

\paragraph{Rarity, curation and budgeted acquisition.}
Mean deep nearest-neighbor distance has been used as a nonparametric out-of-distribution score~\citep{sun2022out}, and we use it as a local-support estimate on the training set itself. Curation by prototype distance or concept-cluster complexity removes examples~\citep{sorscher2022beyond,abbas2024effective,whang2023data}, SMOTE interpolates labeled minority examples~\citep{chawla2002smote}, and DOPING oversamples the low-density edge of a learned latent distribution for anomaly detection with no image generator and no recomputed target~\citep{lim2018doping}, which our top-tail selection rule reproduces under our generator and budget. Active learning spends a finite budget under a learned representation, its core-set form treating acquisition as $k$-center coverage under the greedy farthest-first traversal we use inside the mode window~\citep{sener2018active,gonzalez1985clustering}. ProbCover instead maximizes covered probability mass at a fixed radius~\citep{yehuda2022covering}, and \citet{hacohen2022budget} find dense regions the better target at small budgets.

\paragraph{Generative augmentation and synthetic sources.}
SDEdit adds noise to an image and denoises it under a diffusion prior~\citep{ho2020denoising,rombach2022highresolution,meng2022sdedit}, and that operator is our repair step. It has been applied to few-shot supervised recognition~\citep{trabucco2023effective}, class-conditional generation improves supervised ImageNet accuracy~\citep{azizi2023synthetic}, and synthetic-data utility depends on prompting, filtering, diversity and domain gap~\citep{he2023synthetic,fan2024scaling}. TADA generates faithful variants of the examples a supervised classifier learns slowly~\citep{nguyen2026tada}, a criterion that requires labels. Synthetic imagery has also been a training source for self-supervised features since \citet{ren2018crossdomain}, from procedural noise and image programs~\citep{baradad2021noise,baradad2022procedural} to generator latents, synthetic ImageNet clones and text-to-image views~\citep{jahanian2022generative,sariyildiz2023fake,tian2023stablerep}, none allocating a budget over an observed skewed corpus. DiffAug jointly trains an encoder and a conditional diffusion generator so that generated positives improve contrastive learning~\citep{zang2024diffaug}, learning that generator from the small skewed source it is meant to repair, while \method\ holds a pretrained generator fixed and asks where it should intervene. Industrial data engines close a loop between a deployed model and its training data, diagnosing failures with vision-language and language models and then curating, auto-labeling and verifying~\citep{liang2024aide,worker2022omniverse}. They work from supervised task failures with annotation in the loop and acquire by semantic-cluster occupancy. A frozen generator also brings knowledge from its pretraining data~\citep{li2023dreamteacher,tang2023dift,yu2025repa}. Appendix~\ref{app:prior-art} tabulates the closest methods.

\section{Theoretical Framework}
\label{sec:theory}
Targeted repair is modeled as budgeted allocation over coherent regions of representation space. A repair helps only where it adds support in a region that contributes to downstream error and the generator preserves the content of its anchor, so the policy weighs support deficit against repairability and redundancy. Let $r_k(z;Z)$ be the distance from $z$ to its $k$th nearest neighbor in $Z$ and $A$ a finite set of valid variants generated from $z$. Appendix~\ref{app:proofs} gives complete proofs.

\begin{proposition}[Monotone local support]
\label{prop:local-support}
For any finite $A$, $r_k(z;Z\cup A)\leq r_k(z;Z)$. The inequality is strict when an added point enters the first $k$ neighbors of $z$ and displaces a neighbor at strictly larger distance.
\end{proposition}
Adding points can only introduce smaller candidate distances, so the $k$th order statistic cannot increase.

Partition the representation into regions $j$ with effective support $n_j$ and model the regional contribution to downstream error by
\begin{equation}
\mathcal B(\mathbf m)=\sum_j p_j a_j(n_j+m_j+\kappa)^{-\gamma},
\label{eq:regional-bound}
\end{equation}
where $p_j$ is the transfer-relevant mass of region $j$, $a_j$ its difficulty, $m_j$ the number of valid repairs allocated to it, and $\kappa,\gamma>0$. Write $\Delta_j(m_j)=p_ja_j[(n_j+m_j+\kappa)^{-\gamma}-(n_j+m_j+1+\kappa)^{-\gamma}]$ for the reduction from one more valid repair. None of $p_j$, $a_j$, $\kappa$, $\gamma$ or the regions is instantiated or estimated in the experiments, and Equation~\ref{eq:regional-bound} is a qualitative model in the diminishing-returns form of empirical scaling laws~\citep{sorscher2022beyond}.

\begin{proposition}[Sparse-region marginal value]
\label{prop:sparse-value}
$\Delta_j(m_j)$ is positive and strictly decreasing in $n_j+m_j$. Between two regions of equal transfer mass and difficulty, the region with less effective support has the larger marginal value.
\end{proposition}
The forward difference of the convex decreasing map $x\mapsto x^{-\gamma}$ is positive and decreasing, which gives the claim.

Sparsity alone is not sufficient, because generation fails on corrupt or incoherent anchors. Let $d$ be a sparsity score, $b(d)$ the marginal value of a valid repair at that score, $\rho(d)$ the probability of a faithful variant and $\lambda\geq0$ the cost of an unfaithful one, so the expected repair utility is
\begin{equation}
u(d)=\rho(d)\,b(d)-\bigl(1-\rho(d)\bigr)\lambda.
\label{eq:repair-utility}
\end{equation}

\begin{proposition}[Interior recoverable-support optimum]
\label{prop:interior}
Suppose $b$ and $\rho$ are positive and continuously differentiable on $[d_{\min},d_{\max}]$ with $b'(d)>0$ and $\rho'(d)<0$, and define $g(d)=b'(d)/(b(d)+\lambda)+\rho'(d)/\rho(d)$. If $g$ is strictly decreasing with $g(d_{\min})>0$ and $g(d_{\max})<0$, then $u$ has a unique maximizer $d^\star$ in the interior, characterized by $g(d^\star)=0$.
\end{proposition}
The sign of $u'$ equals the sign of $g$, so the unique sign change of $g$ is the unique maximum.

The monotonicity assumption on $g$ holds for $b(d)=b_0d^{\beta}$ with $\beta\in(0,1]$ and $\rho(d)=\rho_0e^{-\mu d}$ (Proposition~\ref{prop:closed-form}), where $g$ is strictly decreasing for every $\lambda\geq0$ and $d^{\star}=\beta/\mu$ at $\lambda=0$. The optimum moves outward when support value is closer to linear and inward when fidelity decays faster. \method\ uses a trimmed score band and estimates neither parameter. For the repeated case, consider a separable objective $U(\mathbf m)=\sum_jU_j(n_j+m_j)$ with each $U_j$ increasing and discretely concave and a budget of $R$ repairs, $\sum_jm_j=R$.

\begin{proposition}[Adaptive allocation under diminishing returns]
\label{prop:adaptive}
For integer allocations, repeatedly assigning one repair to the region with the largest current marginal gain maximizes $U(\mathbf m)$. A fixed allocation may become suboptimal once a region's marginal gain falls below that of another region.
\end{proposition}
Because every marginal sequence is nonincreasing, the greedy picks are the largest feasible marginals overall. In \method\ the encoder changes after every repair, and the regions and their marginals with it, so the benefit of re-estimation in the non-stationary case is an empirical claim.

\begin{proposition}[Single-stage insensitivity]
\label{prop:one-stage}
Let $\mathbf m$ and $\mathbf m'$ be any two allocations of the same budget of $R$ repairs under Equation~\ref{eq:regional-bound}, computed from one representation and applied without re-estimation. Then
\begin{equation}
\bigl|\mathcal B(\mathbf m)-\mathcal B(\mathbf m')\bigr|\;\leq\;R\,\gamma\max_j p_ja_j\,(n_j+\kappa)^{-\gamma-1},
\end{equation}
which tends to zero as $R/n_{\min}\to0$ for the least-supported region $n_{\min}$.
\end{proposition}
Both allocations subtract at most $R$ marginals from the same baseline, and the mean value theorem bounds every marginal.

A single allocation can therefore differ from any other only in proportion to the budget it adds, and the effect of the acquisition rule accumulates when the allocation is re-estimated over several cycles.

\section{BRIDGE: Balancing Representations by Identifying and Generating Under-represented Data}
\label{sec:method}
\subsection{Loop and scoring}
Let $D^{(0)}$ be an unlabeled source set and $f_\theta$ an SSL encoder. \method\ alternates pretraining and repair over $C$ cycles, with encoder $f_{\theta^{(c)}}$, source set $D^{(c)}$ and embedding set $Z^{(c)}$ at cycle $c$. The encoder is the only learned component, so the loop applies to any objective that yields embeddings. Every cycle continues from the previous encoder with a fresh optimizer schedule and retains all earlier images.

After training on $D^{(c)}$ we embed each image, $z_i^{(c)}=f_{\theta^{(c)}}(x_i)$, normalize to unit length, and compute
\begin{equation}
d_i^{(c)}=\frac{1}{k}\sum_{z\in\mathcal N_k(z_i^{(c)};\,Z^{(c)}\setminus\{z_i^{(c)}\})}\lVert z_i^{(c)}-z\rVert_2^2,
\label{eq:score}
\end{equation}
where $\mathcal N_k(z;Z)$ is the set of $k$ nearest neighbors of $z$ in $Z$, so a large score marks an image in a sparse part of the current representation. We use $k=100$.

\subsection{Sparse and recoverable anchor selection}
\label{sec:mode-window}
\begin{wrapfigure}{r}{0.48\textwidth}
\vspace{-8pt}
\centering
\includegraphics[width=0.47\textwidth]{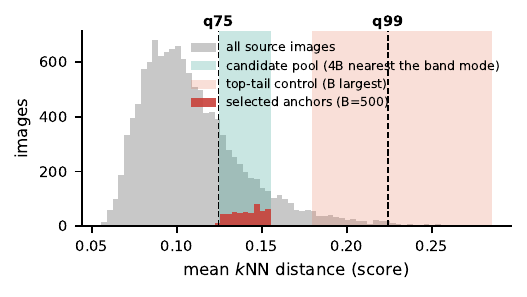}
\vspace{-6pt}
\caption{Score histogram of one ImageNet-100-LT run at cycle 0 with the 75th and 99th percentiles marked. The green pool holds the $4B$ scores nearest the modal bin of the band, which lies at its lower edge, the red region is the top-tail selection rule, and the 500 selected anchors are overlaid.}
\label{fig:mode-window}
\vspace{-6pt}
\end{wrapfigure}
The extreme tail holds artifacts, degenerate crops and images the generator cannot vary faithfully, while uniform sampling spends most of its budget in regions that are already dense. In three steps (Figure~\ref{fig:mode-window}) the selector restricts attention to the band $U^{(c)}=\{i\mid q^{(c)}_{0.75}\leq d_i^{(c)}\leq q^{(c)}_{0.99}\}$, locates the center $\bar d^{(c)}$ of the modal bin of a histogram over that band, forms the pool $P^{(c)}\subseteq U^{(c)}$ of size $\alpha B$ that minimizes $\sum_{i\in P^{(c)}}|d_i^{(c)}-\bar d^{(c)}|$, and selects the anchor set $S^{(c)}$ of size $B$ from that pool by greedy farthest-point sampling in embedding space. In every run the modal bin of the right-skewed band sits at its lower edge, so the candidate pool is the $4B$ scores just above the 75th percentile and the histogram step changes nothing (Figure~\ref{fig:mode-window}). We use $\alpha=4$ and $B=500$.

\subsection{Local generative repair and continued pretraining}
For each anchor we generate $G=5$ variants with a frozen image-to-image diffusion model $g_\phi$, conditioning on the anchor image under an empty text prompt so the repair signal comes from the anchor alone, and set $D^{(c+1)}=D^{(c)}\cup\{g_\phi(x_i,\epsilon_j)\mid i\in S^{(c)},\,j\in[G]\}$. The noise $\epsilon_j$ indexes independent translations of one anchor, so variants keep coarse structure while changing texture, pose or color. We use Stable Diffusion 3 Medium with 20 denoising steps and strength 0.6 at the training crop resolution, and Algorithm~\ref{alg:bridge} gives the loop.

\section{Experimental Setup}
\label{sec:setup}
\paragraph{Sources, targets and protocol.}
We pretrain on five unlabeled source regimes, the long-tailed ImageNet-100-LT~\citep{imagenet-lt}, CIFAR-10-LT and CIFAR-100-LT~\citep{krizhevsky2009learning}, and 10{,}000-image subsets of the web corpora PASS~\citep{pass} and DiffusionDB~\citep{diffusiondb}. Transfer is measured on seven benchmarks, CIFAR-10-LT, CIFAR-100-LT, Stanford Cars~\citep{cars}, FGVC Aircraft~\citep{aircraft}, Oxford Flowers~\citep{flowers}, Oxford-IIIT Pets~\citep{pets} and ImageNet-100-LT, by linear probe and by $k$NN classification on frozen features. Every number is a three-seed mean and standard deviation, paired comparisons share seeds and checkpoints, and ResNet-50 is the default backbone. The baselines train five cycles of 100 epochs on a source that does not grow, while \method\ trains the same epochs on a source that grows from about 10{,}000 to about 20{,}000 images, so it receives about 1.5 times the gradient updates in Tables~\ref{tab:main_transfer_linear} and~\ref{tab:main_transfer_knn}. The cycle ablation and the paired ViT-S experiments match updates exactly. The seed standard deviation of a seven-task average is 0.6 to 1.5 points, so differences below about one point are inside seed variation. Appendix~\ref{app:implementation} gives implementation details.

\paragraph{Long-tailed construction.}
ImageNet-100 is class-balanced, so we construct its long-tailed version. A Pareto profile with exponent $\alpha=2.5$ over the 100 class ranks is rescaled to per-class retention counts between 1{,}280 and 5 images and permuted over the classes. Each image is kept with the retention probability of its class. The balanced source of Table~\ref{tab:main_controlled_imbalance} uses the average of those probabilities for every class, so the two sources have the same expected size and a different class distribution. CIFAR-10-LT and CIFAR-100-LT use their standard configurations.

\paragraph{Compared methods and baselines.}
We compare with two groups of methods. TS and SDCLR change the loss on a fixed long-tailed source and are trained on each of the five sources, with hybrids in the appendix. TADA, AIDE and DiffAug add generated images as \method\ does, so they start from the same ImageNet-100-LT checkpoint per seed and follow the same five cycles, and only the choice or generation of images differs. All ablations use ImageNet-100-LT pretraining, the same downstream suite and three seeds under the five-cycle protocol. The generator ablation includes an oracle that restores 2{,}500 withheld real images per repair, the same count as the generated variants, placing true support at a deficit no label-free method can see.

\section{Results}
\label{sec:results}
\begin{table*}[t]
\centering
\tiny
\setlength{\tabcolsep}{2.6pt}
\renewcommand{\arraystretch}{0.9}
\caption{Imposed imbalance on ImageNet-100 over three seeds. Balancing the classes changes SimCLR by 0.1 points, while \method\ on the long-tailed source adds 11.1 linear-probe and 7.3 $k$NN points.}
\label{tab:main_controlled_imbalance}
\begin{adjustbox}{max width=\textwidth}
\begin{tabular}{l l c c c c c c c c}
\toprule
Protocol & Source & C10 & C100 & Cars & Aircraft & Flowers & Pets & IN100-LT & Avg. \\
\midrule
Linear probe & Balanced & 65.3 {\scriptsize$\pm$ 1.2} & 37.0 {\scriptsize$\pm$ 0.8} & 8.0 {\scriptsize$\pm$ 0.5} & 15.1 {\scriptsize$\pm$ 0.7} & 22.4 {\scriptsize$\pm$ 1.6} & 31.2 {\scriptsize$\pm$ 2.0} & 39.0 {\scriptsize$\pm$ 1.0} & 31.1 {\scriptsize$\pm$ 1.1} \\
 & Purposefully imbalanced & 65.4 {\scriptsize$\pm$ 4.2} & 38.6 {\scriptsize$\pm$ 1.4} & 7.6 {\scriptsize$\pm$ 1.1} & 15.3 {\scriptsize$\pm$ 0.9} & 23.7 {\scriptsize$\pm$ 1.0} & 28.4 {\scriptsize$\pm$ 1.3} & 39.6 {\scriptsize$\pm$ 0.3} & 31.2 {\scriptsize$\pm$ 1.4} \\
\rowcolor{blue!10} & BRIDGE on the imbalanced source & \textbf{76.0 {\scriptsize$\pm$ 0.3}} & \textbf{49.1 {\scriptsize$\pm$ 0.2}} & \textbf{15.2 {\scriptsize$\pm$ 1.2}} & \textbf{22.4 {\scriptsize$\pm$ 0.4}} & \textbf{36.8 {\scriptsize$\pm$ 1.9}} & \textbf{42.9 {\scriptsize$\pm$ 1.0}} & \textbf{53.8 {\scriptsize$\pm$ 0.1}} & \textbf{42.3 {\scriptsize$\pm$ 0.7}} \\
\midrule
$k$NN & Balanced & 63.6 {\scriptsize$\pm$ 0.7} & 32.8 {\scriptsize$\pm$ 0.8} & 4.4 {\scriptsize$\pm$ 0.4} & 9.0 {\scriptsize$\pm$ 0.4} & 22.3 {\scriptsize$\pm$ 0.1} & 17.2 {\scriptsize$\pm$ 0.1} & 34.0 {\scriptsize$\pm$ 1.1} & 26.2 {\scriptsize$\pm$ 0.5} \\
 & Purposefully imbalanced & 63.9 {\scriptsize$\pm$ 0.2} & 33.2 {\scriptsize$\pm$ 0.3} & 4.5 {\scriptsize$\pm$ 0.6} & 9.0 {\scriptsize$\pm$ 0.3} & 22.5 {\scriptsize$\pm$ 0.3} & 16.8 {\scriptsize$\pm$ 0.2} & 34.3 {\scriptsize$\pm$ 0.3} & 26.3 {\scriptsize$\pm$ 0.3} \\
\rowcolor{blue!10} & BRIDGE on the imbalanced source & \textbf{69.0 {\scriptsize$\pm$ 0.5}} & \textbf{37.8 {\scriptsize$\pm$ 0.3}} & \textbf{6.7 {\scriptsize$\pm$ 0.3}} & \textbf{11.7 {\scriptsize$\pm$ 0.2}} & \textbf{36.5 {\scriptsize$\pm$ 0.4}} & \textbf{25.3 {\scriptsize$\pm$ 1.0}} & \textbf{48.1 {\scriptsize$\pm$ 0.6}} & \textbf{33.6 {\scriptsize$\pm$ 0.5}} \\
\bottomrule
\end{tabular}
\end{adjustbox}
\end{table*}

\paragraph{Imposed imbalance on ImageNet-100.} ImageNet is class-balanced by construction, so the first question is what a class imbalance costs once it is imposed on purpose. Two sources are drawn from the same pool, one with the power-law profile of Section~\ref{sec:setup} and one subsampled uniformly at its average retention rate, so they differ only in how images are spread over classes. The two SimCLR runs differ by 0.1 points, 31.1 against 31.2 on the linear-probe seven-task average and 26.2 against 26.3 under $k$NN. \method\ on the imbalanced source reaches 42.3 and 33.6, which is 11.1 and 7.3 points above SimCLR on the same source and a similar margin above the balanced one. The remaining experiments use sources that are long-tailed without our intervention.

\begin{table*}[t]
\centering
\tiny
\setlength{\tabcolsep}{2.6pt}
\renewcommand{\arraystretch}{0.9}
\caption{Linear-probe transfer from five sources to seven tasks, mean $\pm$ standard deviation over three seeds. \method\ has the best average on every source, 2.4 to 10.6 points above SimCLR. Bold marks the best value in a column.}
\label{tab:main_transfer_linear}
\begin{adjustbox}{max width=\textwidth}
\begin{tabular}{l l c c c c c c c c}
\toprule
Source & Method & C10 & C100 & Cars & Aircraft & Flowers & Pets & IN100-LT & Avg. \\
\midrule
ImageNet-100-LT & SimCLR & 73.9 {\scriptsize$\pm$ 0.5} & 46.8 {\scriptsize$\pm$ 0.4} & 13.7 {\scriptsize$\pm$ 0.3} & 19.6 {\scriptsize$\pm$ 0.2} & 35.9 {\scriptsize$\pm$ 7.2} & 38.6 {\scriptsize$\pm$ 1.5} & 50.4 {\scriptsize$\pm$ 0.5} & 39.8 {\scriptsize$\pm$ 1.5} \\
 & TS & 72.5 {\scriptsize$\pm$ 0.4} & 44.0 {\scriptsize$\pm$ 0.5} & 12.5 {\scriptsize$\pm$ 1.2} & 15.0 {\scriptsize$\pm$ 0.1} & 35.7 {\scriptsize$\pm$ 1.8} & 40.6 {\scriptsize$\pm$ 2.2} & \textbf{56.7 {\scriptsize$\pm$ 1.2}} & 39.6 {\scriptsize$\pm$ 1.1} \\
 & SDCLR & 41.1 {\scriptsize$\pm$ 1.7} & 18.0 {\scriptsize$\pm$ 0.6} & 1.4 {\scriptsize$\pm$ 0.3} & 2.3 {\scriptsize$\pm$ 0.5} & 10.8 {\scriptsize$\pm$ 1.7} & 10.4 {\scriptsize$\pm$ 0.3} & 21.9 {\scriptsize$\pm$ 0.6} & 15.1 {\scriptsize$\pm$ 0.8} \\
\rowcolor{blue!10} & BRIDGE (ours) & \textbf{75.8 {\scriptsize$\pm$ 0.2}} & \textbf{49.3 {\scriptsize$\pm$ 0.2}} & \textbf{16.4 {\scriptsize$\pm$ 1.2}} & \textbf{22.4 {\scriptsize$\pm$ 0.4}} & \textbf{36.8 {\scriptsize$\pm$ 1.9}} & \textbf{42.9 {\scriptsize$\pm$ 1.0}} & 53.8 {\scriptsize$\pm$ 0.1} & \textbf{42.5 {\scriptsize$\pm$ 0.7}} \\
\midrule
CIFAR-10-LT & SimCLR & 67.8 {\scriptsize$\pm$ 0.2} & 38.0 {\scriptsize$\pm$ 1.0} & 6.5 {\scriptsize$\pm$ 0.4} & 16.0 {\scriptsize$\pm$ 0.9} & \textbf{19.2 {\scriptsize$\pm$ 1.6}} & 24.0 {\scriptsize$\pm$ 1.3} & 32.0 {\scriptsize$\pm$ 0.7} & 29.1 {\scriptsize$\pm$ 0.9} \\
 & TS & 68.5 {\scriptsize$\pm$ 1.3} & 36.0 {\scriptsize$\pm$ 1.7} & 4.3 {\scriptsize$\pm$ 0.2} & 10.3 {\scriptsize$\pm$ 0.1} & 13.1 {\scriptsize$\pm$ 2.0} & 19.7 {\scriptsize$\pm$ 2.3} & 31.5 {\scriptsize$\pm$ 1.2} & 26.2 {\scriptsize$\pm$ 1.3} \\
 & SDCLR & 39.6 {\scriptsize$\pm$ 1.1} & 15.6 {\scriptsize$\pm$ 0.2} & 1.1 {\scriptsize$\pm$ 0.1} & 2.4 {\scriptsize$\pm$ 0.9} & 12.0 {\scriptsize$\pm$ 1.6} & 9.1 {\scriptsize$\pm$ 1.8} & 16.5 {\scriptsize$\pm$ 1.6} & 13.8 {\scriptsize$\pm$ 1.1} \\
\rowcolor{blue!10} & BRIDGE (ours) & \textbf{78.1 {\scriptsize$\pm$ 0.3}} & \textbf{49.8 {\scriptsize$\pm$ 0.4}} & \textbf{9.2 {\scriptsize$\pm$ 0.8}} & \textbf{17.7 {\scriptsize$\pm$ 1.0}} & 18.2 {\scriptsize$\pm$ 1.1} & \textbf{27.4 {\scriptsize$\pm$ 2.7}} & \textbf{39.0 {\scriptsize$\pm$ 0.6}} & \textbf{34.2 {\scriptsize$\pm$ 1.0}} \\
\midrule
CIFAR-100-LT & SimCLR & 59.0 {\scriptsize$\pm$ 1.5} & 31.1 {\scriptsize$\pm$ 1.6} & 4.1 {\scriptsize$\pm$ 0.3} & 11.2 {\scriptsize$\pm$ 1.0} & 12.6 {\scriptsize$\pm$ 3.4} & 20.1 {\scriptsize$\pm$ 1.0} & 28.7 {\scriptsize$\pm$ 1.3} & 23.8 {\scriptsize$\pm$ 1.4} \\
 & TS & 56.3 {\scriptsize$\pm$ 0.2} & 28.0 {\scriptsize$\pm$ 0.1} & 2.8 {\scriptsize$\pm$ 0.6} & 8.5 {\scriptsize$\pm$ 0.2} & 15.5 {\scriptsize$\pm$ 2.5} & 16.2 {\scriptsize$\pm$ 3.1} & 26.2 {\scriptsize$\pm$ 1.8} & 21.9 {\scriptsize$\pm$ 1.2} \\
 & SDCLR & 37.1 {\scriptsize$\pm$ 0.8} & 15.7 {\scriptsize$\pm$ 0.2} & 1.9 {\scriptsize$\pm$ 0.1} & 2.4 {\scriptsize$\pm$ 0.7} & 9.4 {\scriptsize$\pm$ 1.0} & 9.9 {\scriptsize$\pm$ 1.7} & 16.2 {\scriptsize$\pm$ 0.3} & 13.2 {\scriptsize$\pm$ 0.7} \\
\rowcolor{blue!10} & BRIDGE (ours) & \textbf{74.2 {\scriptsize$\pm$ 0.5}} & \textbf{48.1 {\scriptsize$\pm$ 0.4}} & \textbf{8.3 {\scriptsize$\pm$ 1.0}} & \textbf{18.2 {\scriptsize$\pm$ 0.7}} & \textbf{24.4 {\scriptsize$\pm$ 2.3}} & \textbf{27.6 {\scriptsize$\pm$ 0.7}} & \textbf{39.8 {\scriptsize$\pm$ 1.1}} & \textbf{34.4 {\scriptsize$\pm$ 1.0}} \\
\midrule
PASS-10k & SimCLR & 72.3 {\scriptsize$\pm$ 0.2} & 45.5 {\scriptsize$\pm$ 0.5} & 11.0 {\scriptsize$\pm$ 0.2} & 17.5 {\scriptsize$\pm$ 1.4} & 33.3 {\scriptsize$\pm$ 1.3} & 33.3 {\scriptsize$\pm$ 0.9} & 44.7 {\scriptsize$\pm$ 0.9} & 36.8 {\scriptsize$\pm$ 0.8} \\
 & TS & 71.5 {\scriptsize$\pm$ 0.2} & 43.5 {\scriptsize$\pm$ 0.4} & 10.2 {\scriptsize$\pm$ 1.2} & 17.2 {\scriptsize$\pm$ 0.2} & 33.1 {\scriptsize$\pm$ 2.1} & 36.4 {\scriptsize$\pm$ 1.8} & \textbf{49.4 {\scriptsize$\pm$ 0.4}} & 37.3 {\scriptsize$\pm$ 0.9} \\
 & SDCLR & 42.2 {\scriptsize$\pm$ 1.1} & 17.7 {\scriptsize$\pm$ 0.8} & 1.8 {\scriptsize$\pm$ 0.3} & 2.2 {\scriptsize$\pm$ 0.8} & 10.6 {\scriptsize$\pm$ 1.3} & 8.2 {\scriptsize$\pm$ 1.6} & 20.8 {\scriptsize$\pm$ 0.4} & 14.8 {\scriptsize$\pm$ 0.9} \\
\rowcolor{blue!10} & BRIDGE (ours) & \textbf{73.9 {\scriptsize$\pm$ 0.3}} & \textbf{47.1 {\scriptsize$\pm$ 0.4}} & \textbf{13.6 {\scriptsize$\pm$ 0.7}} & \textbf{20.5 {\scriptsize$\pm$ 0.7}} & \textbf{37.7 {\scriptsize$\pm$ 4.0}} & \textbf{37.1 {\scriptsize$\pm$ 0.7}} & 49.1 {\scriptsize$\pm$ 0.5} & \textbf{39.9 {\scriptsize$\pm$ 1.0}} \\
\midrule
DiffusionDB-10k & SimCLR & 74.4 {\scriptsize$\pm$ 0.5} & 48.6 {\scriptsize$\pm$ 0.2} & 10.9 {\scriptsize$\pm$ 0.7} & 17.6 {\scriptsize$\pm$ 0.8} & 31.2 {\scriptsize$\pm$ 2.4} & 36.1 {\scriptsize$\pm$ 1.3} & 45.3 {\scriptsize$\pm$ 0.5} & 37.7 {\scriptsize$\pm$ 0.9} \\
 & TS & 72.3 {\scriptsize$\pm$ 0.3} & 44.4 {\scriptsize$\pm$ 0.3} & 10.7 {\scriptsize$\pm$ 0.6} & 15.4 {\scriptsize$\pm$ 0.6} & 33.1 {\scriptsize$\pm$ 1.1} & \textbf{39.9 {\scriptsize$\pm$ 2.7}} & 48.8 {\scriptsize$\pm$ 0.2} & 37.8 {\scriptsize$\pm$ 0.8} \\
 & SDCLR & 42.4 {\scriptsize$\pm$ 0.2} & 18.4 {\scriptsize$\pm$ 0.2} & 1.1 {\scriptsize$\pm$ 0.2} & 2.4 {\scriptsize$\pm$ 0.1} & 13.0 {\scriptsize$\pm$ 3.7} & 8.2 {\scriptsize$\pm$ 1.4} & 21.0 {\scriptsize$\pm$ 0.6} & 15.2 {\scriptsize$\pm$ 0.9} \\
\rowcolor{blue!10} & BRIDGE (ours) & \textbf{75.9 {\scriptsize$\pm$ 0.4}} & \textbf{49.9 {\scriptsize$\pm$ 0.5}} & \textbf{12.8 {\scriptsize$\pm$ 1.0}} & \textbf{20.0 {\scriptsize$\pm$ 0.4}} & \textbf{35.7 {\scriptsize$\pm$ 1.1}} & 37.7 {\scriptsize$\pm$ 0.8} & \textbf{49.1 {\scriptsize$\pm$ 0.5}} & \textbf{40.1 {\scriptsize$\pm$ 0.7}} \\
\bottomrule
\end{tabular}
\end{adjustbox}
\end{table*}

\begin{table*}[t]
\centering
\tiny
\setlength{\tabcolsep}{2.6pt}
\renewcommand{\arraystretch}{0.9}
\caption{$k$NN counterpart of Table~\ref{tab:main_transfer_linear}. \method\ again has the best average on every source.}
\label{tab:main_transfer_knn}
\begin{adjustbox}{max width=\textwidth}
\begin{tabular}{l l c c c c c c c c}
\toprule
Source & Method & C10 & C100 & Cars & Aircraft & Flowers & Pets & IN100-LT & Avg. \\
\midrule
ImageNet-100-LT & SimCLR & 68.6 {\scriptsize$\pm$ 0.4} & 37.3 {\scriptsize$\pm$ 0.2} & 6.0 {\scriptsize$\pm$ 0.7} & 11.2 {\scriptsize$\pm$ 0.4} & 32.2 {\scriptsize$\pm$ 0.7} & 23.6 {\scriptsize$\pm$ 0.3} & 43.9 {\scriptsize$\pm$ 0.9} & 31.8 {\scriptsize$\pm$ 0.5} \\
 & TS & 68.4 {\scriptsize$\pm$ 0.3} & 36.5 {\scriptsize$\pm$ 0.1} & 4.7 {\scriptsize$\pm$ 0.2} & 8.3 {\scriptsize$\pm$ 0.5} & 26.6 {\scriptsize$\pm$ 1.0} & 23.6 {\scriptsize$\pm$ 1.7} & \textbf{52.2 {\scriptsize$\pm$ 0.5}} & 31.5 {\scriptsize$\pm$ 0.6} \\
 & SDCLR & 32.5 {\scriptsize$\pm$ 0.4} & 10.9 {\scriptsize$\pm$ 0.2} & 1.0 {\scriptsize$\pm$ 0.3} & 2.1 {\scriptsize$\pm$ 0.1} & 10.7 {\scriptsize$\pm$ 0.7} & 4.7 {\scriptsize$\pm$ 0.2} & 11.2 {\scriptsize$\pm$ 0.2} & 10.4 {\scriptsize$\pm$ 0.3} \\
\rowcolor{blue!10} & BRIDGE (ours) & \textbf{69.0 {\scriptsize$\pm$ 0.5}} & \textbf{37.8 {\scriptsize$\pm$ 0.3}} & \textbf{7.0 {\scriptsize$\pm$ 0.2}} & \textbf{11.7 {\scriptsize$\pm$ 0.2}} & \textbf{36.5 {\scriptsize$\pm$ 0.4}} & \textbf{25.3 {\scriptsize$\pm$ 1.0}} & 48.1 {\scriptsize$\pm$ 0.6} & \textbf{33.6 {\scriptsize$\pm$ 0.4}} \\
\midrule
CIFAR-10-LT & SimCLR & 66.9 {\scriptsize$\pm$ 0.4} & 33.4 {\scriptsize$\pm$ 0.3} & 3.7 {\scriptsize$\pm$ 0.4} & 9.8 {\scriptsize$\pm$ 0.7} & 16.5 {\scriptsize$\pm$ 0.8} & 13.9 {\scriptsize$\pm$ 0.8} & 24.0 {\scriptsize$\pm$ 0.8} & 24.0 {\scriptsize$\pm$ 0.6} \\
 & TS & 66.3 {\scriptsize$\pm$ 0.6} & 27.8 {\scriptsize$\pm$ 0.4} & 2.4 {\scriptsize$\pm$ 0.6} & 6.4 {\scriptsize$\pm$ 0.2} & 10.5 {\scriptsize$\pm$ 0.7} & 11.1 {\scriptsize$\pm$ 0.4} & 22.0 {\scriptsize$\pm$ 0.9} & 20.9 {\scriptsize$\pm$ 0.5} \\
 & SDCLR & 30.1 {\scriptsize$\pm$ 0.3} & 10.0 {\scriptsize$\pm$ 0.3} & 1.7 {\scriptsize$\pm$ 0.3} & 2.2 {\scriptsize$\pm$ 0.2} & 10.1 {\scriptsize$\pm$ 0.7} & 4.4 {\scriptsize$\pm$ 0.4} & 9.8 {\scriptsize$\pm$ 0.3} & 9.7 {\scriptsize$\pm$ 0.4} \\
\rowcolor{blue!10} & BRIDGE (ours) & \textbf{74.8 {\scriptsize$\pm$ 0.4}} & \textbf{41.0 {\scriptsize$\pm$ 0.4}} & \textbf{4.8 {\scriptsize$\pm$ 0.5}} & \textbf{10.0 {\scriptsize$\pm$ 0.2}} & \textbf{24.2 {\scriptsize$\pm$ 0.6}} & \textbf{16.2 {\scriptsize$\pm$ 0.7}} & \textbf{29.7 {\scriptsize$\pm$ 0.3}} & \textbf{28.7 {\scriptsize$\pm$ 0.4}} \\
\midrule
CIFAR-100-LT & SimCLR & 59.3 {\scriptsize$\pm$ 0.8} & 30.1 {\scriptsize$\pm$ 0.9} & 3.0 {\scriptsize$\pm$ 0.3} & 7.4 {\scriptsize$\pm$ 0.4} & 13.7 {\scriptsize$\pm$ 0.2} & 10.8 {\scriptsize$\pm$ 0.4} & 21.8 {\scriptsize$\pm$ 1.1} & 20.9 {\scriptsize$\pm$ 0.6} \\
 & TS & 53.1 {\scriptsize$\pm$ 0.6} & 24.6 {\scriptsize$\pm$ 0.4} & 2.4 {\scriptsize$\pm$ 0.4} & 5.5 {\scriptsize$\pm$ 0.3} & 9.3 {\scriptsize$\pm$ 0.4} & 8.3 {\scriptsize$\pm$ 0.2} & 18.7 {\scriptsize$\pm$ 1.0} & 17.4 {\scriptsize$\pm$ 0.5} \\
 & SDCLR & 25.7 {\scriptsize$\pm$ 1.2} & 8.5 {\scriptsize$\pm$ 0.6} & 0.9 {\scriptsize$\pm$ 0.4} & 1.9 {\scriptsize$\pm$ 0.2} & 7.3 {\scriptsize$\pm$ 0.3} & 3.9 {\scriptsize$\pm$ 0.3} & 8.3 {\scriptsize$\pm$ 1.1} & 8.1 {\scriptsize$\pm$ 0.6} \\
\rowcolor{blue!10} & BRIDGE (ours) & \textbf{69.3 {\scriptsize$\pm$ 0.5}} & \textbf{39.5 {\scriptsize$\pm$ 0.1}} & \textbf{4.8 {\scriptsize$\pm$ 0.6}} & \textbf{8.9 {\scriptsize$\pm$ 0.1}} & \textbf{21.9 {\scriptsize$\pm$ 0.1}} & \textbf{15.5 {\scriptsize$\pm$ 0.1}} & \textbf{29.1 {\scriptsize$\pm$ 0.7}} & \textbf{27.0 {\scriptsize$\pm$ 0.3}} \\
\midrule
PASS-10k & SimCLR & 67.2 {\scriptsize$\pm$ 0.2} & \textbf{37.0 {\scriptsize$\pm$ 0.3}} & 5.4 {\scriptsize$\pm$ 0.3} & 10.3 {\scriptsize$\pm$ 0.5} & 30.4 {\scriptsize$\pm$ 0.0} & 16.7 {\scriptsize$\pm$ 0.5} & 38.3 {\scriptsize$\pm$ 0.3} & 29.3 {\scriptsize$\pm$ 0.3} \\
 & TS & 65.2 {\scriptsize$\pm$ 0.7} & 34.6 {\scriptsize$\pm$ 0.3} & 4.4 {\scriptsize$\pm$ 0.2} & 7.9 {\scriptsize$\pm$ 0.2} & 24.5 {\scriptsize$\pm$ 0.1} & 14.4 {\scriptsize$\pm$ 0.4} & 39.3 {\scriptsize$\pm$ 0.4} & 27.2 {\scriptsize$\pm$ 0.3} \\
 & SDCLR & 32.4 {\scriptsize$\pm$ 1.5} & 11.5 {\scriptsize$\pm$ 0.8} & 1.3 {\scriptsize$\pm$ 0.2} & 2.0 {\scriptsize$\pm$ 0.1} & 8.6 {\scriptsize$\pm$ 0.7} & 5.2 {\scriptsize$\pm$ 0.2} & 11.1 {\scriptsize$\pm$ 0.8} & 10.3 {\scriptsize$\pm$ 0.6} \\
\rowcolor{blue!10} & BRIDGE (ours) & \textbf{67.5 {\scriptsize$\pm$ 0.3}} & 36.9 {\scriptsize$\pm$ 0.3} & \textbf{5.8 {\scriptsize$\pm$ 0.5}} & \textbf{10.7 {\scriptsize$\pm$ 0.4}} & \textbf{33.7 {\scriptsize$\pm$ 0.2}} & \textbf{17.8 {\scriptsize$\pm$ 0.6}} & \textbf{41.4 {\scriptsize$\pm$ 0.8}} & \textbf{30.6 {\scriptsize$\pm$ 0.4}} \\
\midrule
DiffusionDB-10k & SimCLR & 67.3 {\scriptsize$\pm$ 0.1} & 36.2 {\scriptsize$\pm$ 0.4} & 4.3 {\scriptsize$\pm$ 0.2} & 8.8 {\scriptsize$\pm$ 0.1} & 26.8 {\scriptsize$\pm$ 0.4} & 17.3 {\scriptsize$\pm$ 0.1} & 36.7 {\scriptsize$\pm$ 0.8} & 28.2 {\scriptsize$\pm$ 0.3} \\
 & TS & 64.7 {\scriptsize$\pm$ 0.4} & 33.6 {\scriptsize$\pm$ 0.3} & 3.1 {\scriptsize$\pm$ 0.5} & 7.3 {\scriptsize$\pm$ 0.3} & 23.2 {\scriptsize$\pm$ 0.4} & 17.0 {\scriptsize$\pm$ 0.0} & 37.8 {\scriptsize$\pm$ 0.6} & 26.7 {\scriptsize$\pm$ 0.4} \\
 & SDCLR & 32.6 {\scriptsize$\pm$ 1.2} & 10.0 {\scriptsize$\pm$ 0.3} & 1.3 {\scriptsize$\pm$ 0.1} & 2.2 {\scriptsize$\pm$ 0.2} & 6.1 {\scriptsize$\pm$ 0.7} & 4.8 {\scriptsize$\pm$ 0.4} & 10.6 {\scriptsize$\pm$ 0.5} & 9.7 {\scriptsize$\pm$ 0.5} \\
\rowcolor{blue!10} & BRIDGE (ours) & \textbf{68.3 {\scriptsize$\pm$ 0.2}} & \textbf{37.4 {\scriptsize$\pm$ 0.4}} & \textbf{5.4 {\scriptsize$\pm$ 0.3}} & \textbf{9.8 {\scriptsize$\pm$ 0.1}} & \textbf{30.4 {\scriptsize$\pm$ 0.2}} & \textbf{18.6 {\scriptsize$\pm$ 0.2}} & \textbf{39.4 {\scriptsize$\pm$ 1.0}} & \textbf{29.9 {\scriptsize$\pm$ 0.3}} \\
\bottomrule
\end{tabular}
\end{adjustbox}
\end{table*}

\paragraph{Transfer across five source regimes.} \method\ has the best seven-task average in every source regime under both protocols (Tables~\ref{tab:main_transfer_linear} and~\ref{tab:main_transfer_knn}), and against SimCLR and TS it is best in 31 of 35 linear-probe cells and 33 of 35 $k$NN cells. The gain over source-matched SimCLR is above ten linear-probe points under CIFAR-100-LT and above five under CIFAR-10-LT, the two smallest and most skewed sources, and 2.4 to 3.1 points under the three larger sources. TS wins the in-domain target under the ImageNet-100-LT and PASS sources, and SDCLR's pruning schedule was tuned for larger sources, so at 10{,}000 images its averages fall to 13 to 15.

\begin{table*}[t]
\centering
\tiny
\setlength{\tabcolsep}{2.6pt}
\renewcommand{\arraystretch}{0.9}
\caption{Paired runs with and without \method, three seeds. (A) Repair raises the linear-probe average of all four SSL objectives on ViT-S. (B) The gain persists under fine-tuning and segmentation. Bold marks the better run of each pair.}
\label{tab:main_generalization}
\begin{adjustbox}{max width=\textwidth}
\begin{tabular}{l l c c c c c c c c c}
\toprule
\multicolumn{11}{l}{\textit{A. Frozen ViT-S transfer under four objectives}} \\
Objective & Method & C10 & C100 & Cars & Aircraft & Flowers & Pets & IN100-LT & Avg. linear & Avg. $k$NN \\
\midrule
SimCLR & SSL baseline & 60.5 {\scriptsize$\pm$ 0.5} & 35.0 {\scriptsize$\pm$ 0.2} & 7.2 {\scriptsize$\pm$ 0.2} & 10.9 {\scriptsize$\pm$ 0.4} & 32.9 {\scriptsize$\pm$ 2.7} & 25.0 {\scriptsize$\pm$ 1.2} & 42.0 {\scriptsize$\pm$ 1.4} & 30.5 & 30.4 \\
\rowcolor{blue!10} & with \method & \textbf{60.9 {\scriptsize$\pm$ 0.3}} & \textbf{36.3 {\scriptsize$\pm$ 0.1}} & \textbf{8.2 {\scriptsize$\pm$ 0.4}} & \textbf{11.8 {\scriptsize$\pm$ 0.4}} & \textbf{36.6 {\scriptsize$\pm$ 2.9}} & \textbf{29.8 {\scriptsize$\pm$ 3.0}} & \textbf{43.9 {\scriptsize$\pm$ 1.1}} & \textbf{32.5} & \textbf{31.1} \\
MoCo v3 & SSL baseline & 60.0 {\scriptsize$\pm$ 0.4} & 34.5 {\scriptsize$\pm$ 0.0} & 7.6 {\scriptsize$\pm$ 0.2} & 11.1 {\scriptsize$\pm$ 1.3} & 30.3 {\scriptsize$\pm$ 1.4} & 29.3 {\scriptsize$\pm$ 3.1} & 44.8 {\scriptsize$\pm$ 1.2} & 31.1 & 33.0 \\
\rowcolor{blue!10} & with \method & \textbf{62.4 {\scriptsize$\pm$ 0.5}} & \textbf{36.3 {\scriptsize$\pm$ 0.1}} & \textbf{9.6 {\scriptsize$\pm$ 0.3}} & \textbf{13.4 {\scriptsize$\pm$ 1.0}} & \textbf{36.8 {\scriptsize$\pm$ 6.1}} & \textbf{31.2 {\scriptsize$\pm$ 4.0}} & \textbf{47.5 {\scriptsize$\pm$ 0.9}} & \textbf{33.9} & 32.8 \\
DINO & SSL baseline & 58.9 {\scriptsize$\pm$ 0.9} & 33.6 {\scriptsize$\pm$ 1.8} & 8.1 {\scriptsize$\pm$ 0.2} & 11.4 {\scriptsize$\pm$ 1.0} & \textbf{38.1 {\scriptsize$\pm$ 5.2}} & 27.5 {\scriptsize$\pm$ 2.3} & 44.6 {\scriptsize$\pm$ 1.5} & 31.8 & 30.5 \\
\rowcolor{blue!10} & with \method & \textbf{61.3 {\scriptsize$\pm$ 0.1}} & \textbf{35.2 {\scriptsize$\pm$ 0.3}} & \textbf{9.5 {\scriptsize$\pm$ 0.9}} & \textbf{13.3 {\scriptsize$\pm$ 0.6}} & 37.2 {\scriptsize$\pm$ 2.6} & \textbf{34.6 {\scriptsize$\pm$ 3.6}} & \textbf{46.8 {\scriptsize$\pm$ 0.8}} & \textbf{34.0} & \textbf{32.0} \\
MAE & SSL baseline & 37.7 {\scriptsize$\pm$ 3.7} & 13.2 {\scriptsize$\pm$ 1.9} & 1.5 {\scriptsize$\pm$ 0.2} & 2.4 {\scriptsize$\pm$ 1.0} & \textbf{10.5 {\scriptsize$\pm$ 5.9}} & \textbf{10.1 {\scriptsize$\pm$ 3.0}} & 19.9 {\scriptsize$\pm$ 2.1} & 13.6 & 9.6 \\
\rowcolor{blue!10} & with \method & \textbf{41.5 {\scriptsize$\pm$ 2.2}} & \textbf{15.0 {\scriptsize$\pm$ 2.4}} & \textbf{1.6 {\scriptsize$\pm$ 0.4}} & \textbf{3.2 {\scriptsize$\pm$ 0.7}} & 8.5 {\scriptsize$\pm$ 2.3} & 8.6 {\scriptsize$\pm$ 2.2} & \textbf{22.2 {\scriptsize$\pm$ 2.4}} & \textbf{14.4} & \textbf{11.6} \\
\bottomrule
\end{tabular}
\end{adjustbox}

\vspace{3pt}
\begin{adjustbox}{max width=0.7\textwidth}
\begin{tabular}{l l c c c}
\toprule
\multicolumn{5}{l}{\textit{B. End-to-end downstream adaptation, SimCLR ResNet-50}} \\
Evaluation & Setting & SSL baseline & With \method & Paired difference \\
\midrule
Full fine-tuning & IN100-LT, 1\% labels & $14.31\pm0.33$ & $\mathbf{16.29\pm1.11}$ & $+1.97\pm0.78$ \\
Full fine-tuning & IN100-LT, 10\% labels & $21.56\pm0.75$ & $\mathbf{24.72\pm1.75}$ & $+3.17\pm1.66$ \\
Segmentation & PASCAL VOC, mIoU & $12.78\pm0.79$ & $\mathbf{15.20\pm0.86}$ & $+2.42\pm0.27$ \\
\bottomrule
\end{tabular}
\end{adjustbox}
\end{table*}

\paragraph{Pre-training objectives and end-to-end adaptation.} Four objectives run paired on ViT-S under their reference recipes and one repair budget, and \method\ raises the seven-task linear-probe average of each, by 2.0 points for SimCLR, 2.8 for MoCo v3 and 2.2 for DINO, and by 0.8 for MAE, inside seed variation. Under $k$NN the MoCo v3 average is unchanged to slightly lower, 33.0 against 32.8, while the MAE average rises 2.0 points. Of the 14 measurements per objective, 12 or 13 improve, and fine-grained targets gain most, up to 7.1 points on Pets for DINO. The same ResNet-50 representations improve after end-to-end adaptation, by 2.0 and 3.2 points of fine-tuning accuracy with 1 and 10 percent of the labels and by 2.4 mIoU on PASCAL VOC segmentation.

\begin{table*}[t]
\centering
\tiny
\setlength{\tabcolsep}{2.6pt}
\renewcommand{\arraystretch}{0.9}
\caption{Prior methods on ImageNet-100-LT with one source checkpoint per seed and three seeds. Averages cover both probes, and the SimCLR and \method\ rows repeat Tables~\ref{tab:main_transfer_linear} and~\ref{tab:main_transfer_knn}.}
\label{tab:main_prior}
\begin{adjustbox}{max width=\textwidth}
\begin{tabular}{l c c c c c c c c c}
\toprule
Method & C10 & C100 & Cars & Aircraft & Flowers & Pets & IN100-LT & Avg. & $k$NN Avg. \\
\midrule
SimCLR & 73.9 {\scriptsize$\pm$ 0.5} & 46.8 {\scriptsize$\pm$ 0.4} & 13.7 {\scriptsize$\pm$ 0.3} & 19.6 {\scriptsize$\pm$ 0.2} & 35.9 {\scriptsize$\pm$ 7.2} & 38.6 {\scriptsize$\pm$ 1.5} & 50.4 {\scriptsize$\pm$ 0.5} & 39.8 {\scriptsize$\pm$ 1.5} & 31.8 {\scriptsize$\pm$ 0.5} \\
DiffAug & 35.5 {\scriptsize$\pm$ 5.7} & 10.2 {\scriptsize$\pm$ 3.3} & 1.1 {\scriptsize$\pm$ 0.2} & 2.7 {\scriptsize$\pm$ 0.1} & 3.1 {\scriptsize$\pm$ 2.1} & 8.2 {\scriptsize$\pm$ 0.7} & 18.6 {\scriptsize$\pm$ 1.5} & 11.3 {\scriptsize$\pm$ 1.6} & 22.0 {\scriptsize$\pm$ 0.2} \\
DiffAug + \method & 36.1 {\scriptsize$\pm$ 4.0} & 10.3 {\scriptsize$\pm$ 2.4} & 1.2 {\scriptsize$\pm$ 0.1} & 2.6 {\scriptsize$\pm$ 0.2} & 1.1 {\scriptsize$\pm$ 1.4} & 7.5 {\scriptsize$\pm$ 2.1} & 19.0 {\scriptsize$\pm$ 1.1} & 11.1 {\scriptsize$\pm$ 1.4} & 22.3 {\scriptsize$\pm$ 0.2} \\
TADA (SSL loss) & 73.9 {\scriptsize$\pm$ 1.5} & 45.0 {\scriptsize$\pm$ 2.2} & 14.4 {\scriptsize$\pm$ 0.8} & 15.3 {\scriptsize$\pm$ 1.4} & \textbf{39.2 {\scriptsize$\pm$ 2.4}} & 40.1 {\scriptsize$\pm$ 1.4} & 50.2 {\scriptsize$\pm$ 1.0} & 39.7 {\scriptsize$\pm$ 1.0} & 31.7 {\scriptsize$\pm$ 0.8} \\
AIDE (CLIP clusters, BLIP captions) & 72.9 {\scriptsize$\pm$ 0.4} & 44.8 {\scriptsize$\pm$ 1.6} & 14.8 {\scriptsize$\pm$ 0.0} & 17.0 {\scriptsize$\pm$ 0.9} & 37.7 {\scriptsize$\pm$ 2.5} & 40.8 {\scriptsize$\pm$ 3.3} & 49.7 {\scriptsize$\pm$ 0.9} & 39.7 {\scriptsize$\pm$ 0.5} & 32.1 {\scriptsize$\pm$ 0.3} \\
\midrule
\rowcolor{blue!10} \method\ (ours) & \textbf{75.8 {\scriptsize$\pm$ 0.2}} & \textbf{49.3 {\scriptsize$\pm$ 0.2}} & \textbf{16.4 {\scriptsize$\pm$ 1.2}} & \textbf{22.4 {\scriptsize$\pm$ 0.4}} & 36.8 {\scriptsize$\pm$ 1.9} & \textbf{42.9 {\scriptsize$\pm$ 1.0}} & \textbf{53.8 {\scriptsize$\pm$ 0.1}} & \textbf{42.5 {\scriptsize$\pm$ 0.7}} & \textbf{33.6 {\scriptsize$\pm$ 0.4}} \\
\bottomrule
\end{tabular}
\end{adjustbox}
\end{table*}

\paragraph{Prior methods.} Table~\ref{tab:main_prior} compares \method\ with three methods that also add generated images. \method\ has the highest average under both probes, 2.8 linear-probe points and 1.5 $k$NN points above the best prior method, and the highest accuracy on six of the seven targets. The exception is Flowers, where TADA reaches 39.2 against 36.8. No prior method raises either average by more than 0.3 points over SimCLR. TADA generates variants of the examples a supervised classifier learns slowly. Without labels, we select the 500 images with the highest InfoNCE loss under the current encoder and repair them with the image-to-image step of \method. AIDE samples images inversely to the size of their cluster in CLIP ViT-B/32 feature space, captions them with BLIP and generates new images from the caption alone. Both use the same generator and image budget as \method, so the gap comes from which images are chosen and how they are generated. DiffAug trains its own diffusion generator jointly with the encoder and reaches 11.3, and adding \method\ repair to it changes the average by less than half a point.

\subsection{Ablations}
\label{sec:ablations}
\begin{table*}[t]
\centering
\tiny
\setlength{\tabcolsep}{2.6pt}
\renewcommand{\arraystretch}{0.9}
\caption{Ablations on ImageNet-100-LT, linear probe, three seeds, defaults shaded. Generated repair matches the real-image oracle, every band window beats uniform and top-tail selection, and five cycles beat two and ten.}
\label{tab:main_ablation_combined}
\begin{adjustbox}{max width=\textwidth}
\begin{tabular}{l l c c c c c c c c c}
\toprule
Group & Setting & C10 & C100 & Cars & Aircraft & Flowers & Pets & IN100-LT & Avg. \\
\midrule
\rowcolor{blue!10} Generator & Stable Diffusion 3 (default) & \textbf{75.0 {\scriptsize$\pm$ 0.3}} & \textbf{48.5 {\scriptsize$\pm$ 0.2}} & \textbf{16.3 {\scriptsize$\pm$ 0.8}} & \textbf{20.7 {\scriptsize$\pm$ 0.4}} & 35.7 {\scriptsize$\pm$ 0.9} & 41.8 {\scriptsize$\pm$ 0.8} & \textbf{52.4 {\scriptsize$\pm$ 0.7}} & \textbf{41.5 {\scriptsize$\pm$ 0.6}} \\
 & FLUX & 70.7 {\scriptsize$\pm$ 0.4} & 42.1 {\scriptsize$\pm$ 1.4} & 10.3 {\scriptsize$\pm$ 1.3} & 17.5 {\scriptsize$\pm$ 0.9} & 29.4 {\scriptsize$\pm$ 4.6} & 31.5 {\scriptsize$\pm$ 2.9} & 43.7 {\scriptsize$\pm$ 1.4} & 35.0 {\scriptsize$\pm$ 1.9} \\
 & Oracle (withheld real images restored) & 74.8 {\scriptsize$\pm$ 0.4} & 47.8 {\scriptsize$\pm$ 0.7} & 13.1 {\scriptsize$\pm$ 0.1} & 20.5 {\scriptsize$\pm$ 0.5} & 40.3 {\scriptsize$\pm$ 6.1} & 38.1 {\scriptsize$\pm$ 1.5} & 50.3 {\scriptsize$\pm$ 0.4} & 40.7 {\scriptsize$\pm$ 1.4} \\
 & BLIP caption to SD3 text-to-image & 72.6 {\scriptsize$\pm$ 1.0} & 45.1 {\scriptsize$\pm$ 1.2} & 14.3 {\scriptsize$\pm$ 0.3} & 17.4 {\scriptsize$\pm$ 0.6} & \textbf{35.9 {\scriptsize$\pm$ 3.7}} & \textbf{43.3 {\scriptsize$\pm$ 2.5}} & 49.2 {\scriptsize$\pm$ 0.0} & 39.7 {\scriptsize$\pm$ 0.1} \\
 & SD3 image-to-image with BLIP caption as prompt & 70.7 {\scriptsize$\pm$ 1.0} & 41.5 {\scriptsize$\pm$ 1.2} & 12.7 {\scriptsize$\pm$ 0.1} & 14.5 {\scriptsize$\pm$ 1.1} & \textbf{35.9 {\scriptsize$\pm$ 5.2}} & 39.7 {\scriptsize$\pm$ 2.4} & 47.4 {\scriptsize$\pm$ 0.4} & 37.5 {\scriptsize$\pm$ 0.8} \\
\midrule
Distillation & SD3 latent regression, no generated images & \textbf{70.7 {\scriptsize$\pm$ 1.1}} & \textbf{42.3 {\scriptsize$\pm$ 1.1}} & \textbf{13.2 {\scriptsize$\pm$ 0.9}} & \textbf{14.6 {\scriptsize$\pm$ 1.6}} & 33.8 {\scriptsize$\pm$ 4.8} & \textbf{39.9 {\scriptsize$\pm$ 2.1}} & 47.1 {\scriptsize$\pm$ 0.5} & \textbf{37.4 {\scriptsize$\pm$ 1.5}} \\
 & SD3 latent regression and \method & 69.6 {\scriptsize$\pm$ 0.4} & 40.9 {\scriptsize$\pm$ 0.6} & 13.0 {\scriptsize$\pm$ 0.1} & 14.1 {\scriptsize$\pm$ 0.1} & \textbf{35.3 {\scriptsize$\pm$ 1.8}} & 38.9 {\scriptsize$\pm$ 6.5} & \textbf{47.8 {\scriptsize$\pm$ 0.1}} & 37.1 {\scriptsize$\pm$ 1.1} \\
\midrule
Selection & Mode window, lower cutoff p1 & 75.6 {\scriptsize$\pm$ 0.1} & 49.2 {\scriptsize$\pm$ 0.2} & \textbf{16.9 {\scriptsize$\pm$ 0.7}} & 21.9 {\scriptsize$\pm$ 0.4} & 34.6 {\scriptsize$\pm$ 2.9} & 43.5 {\scriptsize$\pm$ 0.7} & 52.8 {\scriptsize$\pm$ 0.1} & 42.1 {\scriptsize$\pm$ 0.7} \\
 & Mode window, lower cutoff p50 & 75.5 {\scriptsize$\pm$ 0.1} & \textbf{49.3 {\scriptsize$\pm$ 0.4}} & 15.2 {\scriptsize$\pm$ 0.3} & \textbf{22.4 {\scriptsize$\pm$ 0.6}} & 35.5 {\scriptsize$\pm$ 1.1} & \textbf{43.8 {\scriptsize$\pm$ 3.1}} & 53.4 {\scriptsize$\pm$ 0.9} & 42.2 {\scriptsize$\pm$ 0.9} \\
\rowcolor{blue!10}  & Mode window, lower cutoff p75 (default) & \textbf{76.0 {\scriptsize$\pm$ 0.3}} & 49.1 {\scriptsize$\pm$ 0.2} & 15.2 {\scriptsize$\pm$ 1.2} & \textbf{22.4 {\scriptsize$\pm$ 0.4}} & \textbf{36.8 {\scriptsize$\pm$ 1.9}} & 42.9 {\scriptsize$\pm$ 1.0} & \textbf{53.8 {\scriptsize$\pm$ 0.1}} & \textbf{42.3 {\scriptsize$\pm$ 0.7}} \\
 & Top tail & 70.8 {\scriptsize$\pm$ 0.5} & 43.1 {\scriptsize$\pm$ 1.0} & 10.0 {\scriptsize$\pm$ 1.7} & 15.9 {\scriptsize$\pm$ 1.3} & 27.7 {\scriptsize$\pm$ 1.0} & 30.5 {\scriptsize$\pm$ 1.2} & 42.2 {\scriptsize$\pm$ 1.4} & 34.3 {\scriptsize$\pm$ 1.2} \\
 & Uniform & 71.5 {\scriptsize$\pm$ 0.2} & 43.8 {\scriptsize$\pm$ 0.7} & 9.3 {\scriptsize$\pm$ 0.9} & 17.5 {\scriptsize$\pm$ 1.2} & 32.9 {\scriptsize$\pm$ 0.4} & 31.7 {\scriptsize$\pm$ 3.9} & 42.0 {\scriptsize$\pm$ 0.9} & 35.5 {\scriptsize$\pm$ 1.2} \\
 & Inverse cluster occupancy & 69.9 {\scriptsize$\pm$ 1.0} & 40.9 {\scriptsize$\pm$ 1.9} & 15.3 {\scriptsize$\pm$ 0.6} & 16.5 {\scriptsize$\pm$ 0.3} & 33.6 {\scriptsize$\pm$ 8.4} & 39.5 {\scriptsize$\pm$ 2.7} & 49.4 {\scriptsize$\pm$ 0.6} & 37.9 {\scriptsize$\pm$ 0.6} \\
\midrule
Cycles & 2 cycles & 68.9 {\scriptsize$\pm$ 0.6} & 41.6 {\scriptsize$\pm$ 1.3} & 10.6 {\scriptsize$\pm$ 1.2} & 16.4 {\scriptsize$\pm$ 0.4} & 30.7 {\scriptsize$\pm$ 4.5} & 33.2 {\scriptsize$\pm$ 0.9} & 43.3 {\scriptsize$\pm$ 0.6} & 35.0 {\scriptsize$\pm$ 1.4} \\
\rowcolor{blue!10}  & 5 cycles (default) & \textbf{76.0 {\scriptsize$\pm$ 0.3}} & \textbf{49.1 {\scriptsize$\pm$ 0.2}} & \textbf{15.2 {\scriptsize$\pm$ 1.2}} & \textbf{22.4 {\scriptsize$\pm$ 0.4}} & \textbf{36.8 {\scriptsize$\pm$ 1.9}} & \textbf{42.9 {\scriptsize$\pm$ 1.0}} & \textbf{53.8 {\scriptsize$\pm$ 0.1}} & \textbf{42.3 {\scriptsize$\pm$ 0.7}} \\
 & 10 cycles & 70.5 {\scriptsize$\pm$ 1.2} & 43.8 {\scriptsize$\pm$ 0.6} & 9.2 {\scriptsize$\pm$ 0.9} & 17.2 {\scriptsize$\pm$ 1.6} & 24.8 {\scriptsize$\pm$ 5.2} & 31.3 {\scriptsize$\pm$ 1.7} & 41.9 {\scriptsize$\pm$ 0.6} & 34.1 {\scriptsize$\pm$ 1.7} \\
\midrule
Architecture & ResNet-18 & 68.8 {\scriptsize$\pm$ 0.1} & 41.7 {\scriptsize$\pm$ 0.2} & 13.5 {\scriptsize$\pm$ 0.3} & 19.7 {\scriptsize$\pm$ 0.6} & 38.7 {\scriptsize$\pm$ 3.4} & 38.4 {\scriptsize$\pm$ 2.6} & 50.2 {\scriptsize$\pm$ 0.2} & 38.7 {\scriptsize$\pm$ 1.1} \\
\rowcolor{blue!10}  & ResNet-50 (default) & \textbf{76.0 {\scriptsize$\pm$ 0.3}} & \textbf{49.1 {\scriptsize$\pm$ 0.2}} & \textbf{15.2 {\scriptsize$\pm$ 1.2}} & \textbf{22.4 {\scriptsize$\pm$ 0.4}} & 36.8 {\scriptsize$\pm$ 1.9} & \textbf{42.9 {\scriptsize$\pm$ 1.0}} & \textbf{53.8 {\scriptsize$\pm$ 0.1}} & \textbf{42.3 {\scriptsize$\pm$ 0.7}} \\
 & ViT-S & 62.7 {\scriptsize$\pm$ 0.4} & 37.6 {\scriptsize$\pm$ 0.6} & 8.8 {\scriptsize$\pm$ 0.5} & 12.2 {\scriptsize$\pm$ 0.2} & 35.1 {\scriptsize$\pm$ 4.2} & 30.2 {\scriptsize$\pm$ 1.6} & 42.5 {\scriptsize$\pm$ 1.2} & 32.7 {\scriptsize$\pm$ 1.3} \\
 & ViT-B & 66.0 {\scriptsize$\pm$ 2.3} & 40.6 {\scriptsize$\pm$ 2.3} & 8.3 {\scriptsize$\pm$ 0.5} & 14.3 {\scriptsize$\pm$ 0.8} & \textbf{38.8 {\scriptsize$\pm$ 2.6}} & 31.3 {\scriptsize$\pm$ 2.1} & 44.5 {\scriptsize$\pm$ 1.7} & 34.8 {\scriptsize$\pm$ 1.8} \\
\bottomrule
\end{tabular}
\end{adjustbox}
\end{table*}

\begin{table*}[t]
\centering
\tiny
\setlength{\tabcolsep}{2.6pt}
\renewcommand{\arraystretch}{0.9}
\caption{$k$NN counterpart of Table~\ref{tab:main_ablation_combined}.}
\label{tab:main_ablation_combined_knn}
\begin{adjustbox}{max width=\textwidth}
\begin{tabular}{l l c c c c c c c c c}
\toprule
Group & Setting & C10 & C100 & Cars & Aircraft & Flowers & Pets & IN100-LT & Avg. \\
\midrule
Generator & Stable Diffusion 3 (default) & 68.6 {\scriptsize$\pm$ 0.1} & 37.2 {\scriptsize$\pm$ 0.2} & 6.4 {\scriptsize$\pm$ 0.6} & 11.5 {\scriptsize$\pm$ 0.2} & \textbf{35.2 {\scriptsize$\pm$ 0.3}} & \textbf{24.3 {\scriptsize$\pm$ 0.6}} & \textbf{46.8 {\scriptsize$\pm$ 0.7}} & \textbf{32.9 {\scriptsize$\pm$ 0.4}} \\
 & FLUX & 65.8 {\scriptsize$\pm$ 1.0} & 35.4 {\scriptsize$\pm$ 0.8} & 5.7 {\scriptsize$\pm$ 0.7} & 9.4 {\scriptsize$\pm$ 0.9} & 27.7 {\scriptsize$\pm$ 1.0} & 18.8 {\scriptsize$\pm$ 1.3} & 38.7 {\scriptsize$\pm$ 1.2} & 28.8 {\scriptsize$\pm$ 1.0} \\
 & Oracle (withheld real images restored) & 68.8 {\scriptsize$\pm$ 0.2} & 37.7 {\scriptsize$\pm$ 0.3} & 6.6 {\scriptsize$\pm$ 0.5} & 11.4 {\scriptsize$\pm$ 0.5} & 32.7 {\scriptsize$\pm$ 0.1} & 23.6 {\scriptsize$\pm$ 0.6} & 44.4 {\scriptsize$\pm$ 0.4} & 32.2 {\scriptsize$\pm$ 0.4} \\
 & BLIP caption to SD3 text-to-image & \textbf{70.1 {\scriptsize$\pm$ 0.4}} & \textbf{39.6 {\scriptsize$\pm$ 0.8}} & 7.2 {\scriptsize$\pm$ 0.5} & \textbf{11.7 {\scriptsize$\pm$ 0.4}} & 31.9 {\scriptsize$\pm$ 1.6} & 21.1 {\scriptsize$\pm$ 0.9} & 41.6 {\scriptsize$\pm$ 0.9} & 31.9 {\scriptsize$\pm$ 0.5} \\
 & SD3 image-to-image with BLIP caption as prompt & 69.6 {\scriptsize$\pm$ 0.1} & 39.1 {\scriptsize$\pm$ 0.6} & \textbf{7.4 {\scriptsize$\pm$ 0.3}} & \textbf{11.7 {\scriptsize$\pm$ 0.2}} & 31.5 {\scriptsize$\pm$ 0.8} & 20.8 {\scriptsize$\pm$ 0.4} & 41.2 {\scriptsize$\pm$ 0.6} & 31.6 {\scriptsize$\pm$ 0.3} \\
\midrule
Distillation & SD3 latent regression, no generated images & \textbf{70.3 {\scriptsize$\pm$ 0.2}} & \textbf{39.3 {\scriptsize$\pm$ 0.2}} & 7.2 {\scriptsize$\pm$ 0.3} & 11.3 {\scriptsize$\pm$ 0.4} & 30.5 {\scriptsize$\pm$ 0.6} & \textbf{21.8 {\scriptsize$\pm$ 0.7}} & \textbf{41.1 {\scriptsize$\pm$ 0.3}} & \textbf{31.7 {\scriptsize$\pm$ 0.1}} \\
 & SD3 latent regression and \method & 68.6 {\scriptsize$\pm$ 0.7} & 38.6 {\scriptsize$\pm$ 0.4} & \textbf{7.5 {\scriptsize$\pm$ 0.6}} & \textbf{11.6 {\scriptsize$\pm$ 0.3}} & \textbf{31.0 {\scriptsize$\pm$ 0.9}} & 20.8 {\scriptsize$\pm$ 1.0} & 40.4 {\scriptsize$\pm$ 0.2} & 31.2 {\scriptsize$\pm$ 0.5} \\
\midrule
Selection & Mode window, lower cutoff p1 & \textbf{69.2 {\scriptsize$\pm$ 0.3}} & 37.7 {\scriptsize$\pm$ 0.2} & 6.6 {\scriptsize$\pm$ 0.2} & 11.4 {\scriptsize$\pm$ 0.2} & 35.9 {\scriptsize$\pm$ 0.2} & 25.0 {\scriptsize$\pm$ 0.5} & 48.3 {\scriptsize$\pm$ 0.3} & 33.4 {\scriptsize$\pm$ 0.3} \\
 & Mode window, lower cutoff p50 & 68.7 {\scriptsize$\pm$ 0.1} & 38.0 {\scriptsize$\pm$ 0.2} & 7.3 {\scriptsize$\pm$ 0.2} & 11.3 {\scriptsize$\pm$ 0.2} & \textbf{36.5 {\scriptsize$\pm$ 0.1}} & 24.8 {\scriptsize$\pm$ 0.8} & \textbf{48.5 {\scriptsize$\pm$ 0.4}} & \textbf{33.6 {\scriptsize$\pm$ 0.3}} \\
 & Mode window, lower cutoff p75 (default) & 69.0 {\scriptsize$\pm$ 0.5} & 37.8 {\scriptsize$\pm$ 0.3} & 6.7 {\scriptsize$\pm$ 0.3} & 11.7 {\scriptsize$\pm$ 0.2} & \textbf{36.5 {\scriptsize$\pm$ 0.4}} & \textbf{25.3 {\scriptsize$\pm$ 1.0}} & 48.1 {\scriptsize$\pm$ 0.6} & \textbf{33.6 {\scriptsize$\pm$ 0.5}} \\
 & Top tail & 65.7 {\scriptsize$\pm$ 0.8} & 34.4 {\scriptsize$\pm$ 0.2} & 4.1 {\scriptsize$\pm$ 0.4} & 9.6 {\scriptsize$\pm$ 0.2} & 23.9 {\scriptsize$\pm$ 0.9} & 18.1 {\scriptsize$\pm$ 0.8} & 35.4 {\scriptsize$\pm$ 1.2} & 27.3 {\scriptsize$\pm$ 0.6} \\
 & Uniform & 66.1 {\scriptsize$\pm$ 1.0} & 35.7 {\scriptsize$\pm$ 0.3} & 4.9 {\scriptsize$\pm$ 0.1} & 10.2 {\scriptsize$\pm$ 0.7} & 25.3 {\scriptsize$\pm$ 0.4} & 19.0 {\scriptsize$\pm$ 0.8} & 37.4 {\scriptsize$\pm$ 0.6} & 28.4 {\scriptsize$\pm$ 0.6} \\
 & Inverse cluster occupancy & 68.7 {\scriptsize$\pm$ 0.3} & \textbf{38.6 {\scriptsize$\pm$ 0.2}} & \textbf{7.7 {\scriptsize$\pm$ 0.2}} & \textbf{11.8 {\scriptsize$\pm$ 0.3}} & 32.1 {\scriptsize$\pm$ 0.9} & 21.4 {\scriptsize$\pm$ 0.6} & 41.2 {\scriptsize$\pm$ 0.4} & 31.6 {\scriptsize$\pm$ 0.2} \\
\midrule
Cycles & 2 cycles & 65.3 {\scriptsize$\pm$ 0.4} & 34.2 {\scriptsize$\pm$ 0.5} & 5.0 {\scriptsize$\pm$ 0.7} & 10.0 {\scriptsize$\pm$ 0.2} & 26.1 {\scriptsize$\pm$ 0.8} & 18.7 {\scriptsize$\pm$ 0.6} & 35.9 {\scriptsize$\pm$ 0.3} & 27.9 {\scriptsize$\pm$ 0.5} \\
 & 5 cycles (default) & \textbf{69.0 {\scriptsize$\pm$ 0.5}} & \textbf{37.8 {\scriptsize$\pm$ 0.3}} & \textbf{6.7 {\scriptsize$\pm$ 0.3}} & \textbf{11.7 {\scriptsize$\pm$ 0.2}} & \textbf{36.5 {\scriptsize$\pm$ 0.4}} & \textbf{25.3 {\scriptsize$\pm$ 1.0}} & \textbf{48.1 {\scriptsize$\pm$ 0.6}} & \textbf{33.6 {\scriptsize$\pm$ 0.5}} \\
 & 10 cycles & 66.4 {\scriptsize$\pm$ 0.5} & 35.4 {\scriptsize$\pm$ 0.3} & 4.8 {\scriptsize$\pm$ 0.3} & 10.4 {\scriptsize$\pm$ 0.6} & 25.2 {\scriptsize$\pm$ 0.8} & 18.6 {\scriptsize$\pm$ 0.9} & 36.0 {\scriptsize$\pm$ 0.7} & 28.1 {\scriptsize$\pm$ 0.6} \\
\midrule
Architecture & ResNet-18 & 66.3 {\scriptsize$\pm$ 0.2} & 35.7 {\scriptsize$\pm$ 0.3} & 5.9 {\scriptsize$\pm$ 0.2} & 11.1 {\scriptsize$\pm$ 0.4} & 33.5 {\scriptsize$\pm$ 0.1} & 23.3 {\scriptsize$\pm$ 0.5} & 44.5 {\scriptsize$\pm$ 0.7} & 31.5 {\scriptsize$\pm$ 0.4} \\
 & ResNet-50 (default) & 69.0 {\scriptsize$\pm$ 0.5} & 37.8 {\scriptsize$\pm$ 0.3} & \textbf{6.7 {\scriptsize$\pm$ 0.3}} & \textbf{11.7 {\scriptsize$\pm$ 0.2}} & 36.5 {\scriptsize$\pm$ 0.4} & \textbf{25.3 {\scriptsize$\pm$ 1.0}} & \textbf{48.1 {\scriptsize$\pm$ 0.6}} & \textbf{33.6 {\scriptsize$\pm$ 0.5}} \\
 & ViT-S & 67.3 {\scriptsize$\pm$ 0.4} & 37.5 {\scriptsize$\pm$ 0.3} & 5.3 {\scriptsize$\pm$ 0.2} & 7.6 {\scriptsize$\pm$ 0.1} & 36.2 {\scriptsize$\pm$ 0.7} & 20.1 {\scriptsize$\pm$ 1.0} & 42.1 {\scriptsize$\pm$ 1.0} & 30.9 {\scriptsize$\pm$ 0.5} \\
 & ViT-B & \textbf{69.3 {\scriptsize$\pm$ 0.8}} & \textbf{40.1 {\scriptsize$\pm$ 0.8}} & 6.0 {\scriptsize$\pm$ 0.2} & 8.3 {\scriptsize$\pm$ 0.2} & \textbf{39.8 {\scriptsize$\pm$ 0.8}} & 21.9 {\scriptsize$\pm$ 1.4} & 45.1 {\scriptsize$\pm$ 0.7} & 32.9 {\scriptsize$\pm$ 0.7} \\
\bottomrule
\end{tabular}
\end{adjustbox}
\end{table*}

\paragraph{Generator.} Generated repair with Stable Diffusion 3 matches the oracle of Section~\ref{sec:setup}, 41.5 against 40.7. With FLUX at six sampling steps and guidance 2.5, the average falls to 35.0, 6.5 points below Stable Diffusion 3. With the BLIP captions of AIDE at the anchors of \method, generating from the caption alone reaches 39.7 and passing it as the prompt of the image-to-image step reaches 37.5.

\paragraph{Selection.} The three mode windows with lower cutoffs at the 1st, 50th and 75th percentile reach 42.1 to 42.3 and are not separated, while uniform acquisition reaches $35.5\pm1.2$ and the top tail $34.3\pm1.2$ (Welch t-tests with Holm adjustment, $p\leq.011$ for every window against both). This agrees with Proposition~\ref{prop:interior}, where excluding the extreme tail raises the objective, and leaves the lower boundary of the band unresolved. Each window changes trimming and diversification at once, and the 6.6 to 8.0 points include both effects. The inverse-occupancy rule of AIDE on $\sqrt{N}$ spherical $k$-means clusters of the encoder features reaches 37.9.

\paragraph{Distillation.} A linear head on the encoder regresses the averaged Stable Diffusion 3 VAE latent of each training image, with a cosine loss of weight 0.1 added to the SimCLR loss. The average is 37.4 without generated images and 37.1 with \method\ repair, 5.2 points below \method\ alone, so the gain of \method\ comes from the generated images at the selected anchors.

\paragraph{Cycles and backbone.} At fixed total repair volume and optimizer budget, five cycles improve the seven-task average over two by 7.3 points and remain 8.2 points ahead of ten. Two cycles spend most of the budget before the representation has changed, and ten leave only 50 epochs of training after each repair. Among the tested backbones, ResNet-50 has the highest average at this source size.

\subsection{Representation geometry}
\label{sec:geometry}
\begin{figure}[t]
\centering
\begin{minipage}[t]{0.42\linewidth}
\centering
\includegraphics[width=\linewidth]{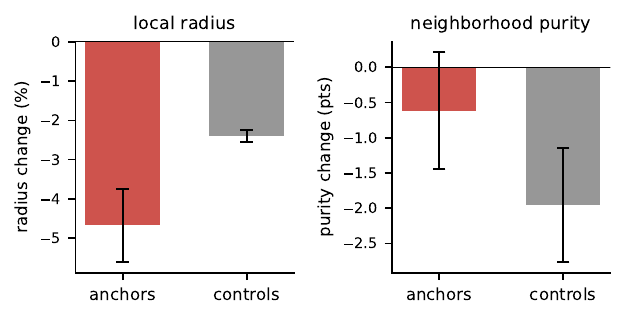}
\caption{With the encoder frozen, generated images shrink the neighborhoods of their own anchors about twice as much as those of matched unselected images. Seed means with one standard deviation.}
\label{fig:fixed-support}
\end{minipage}\hfill
\begin{minipage}[t]{0.50\linewidth}
\centering
\includegraphics[width=\linewidth]{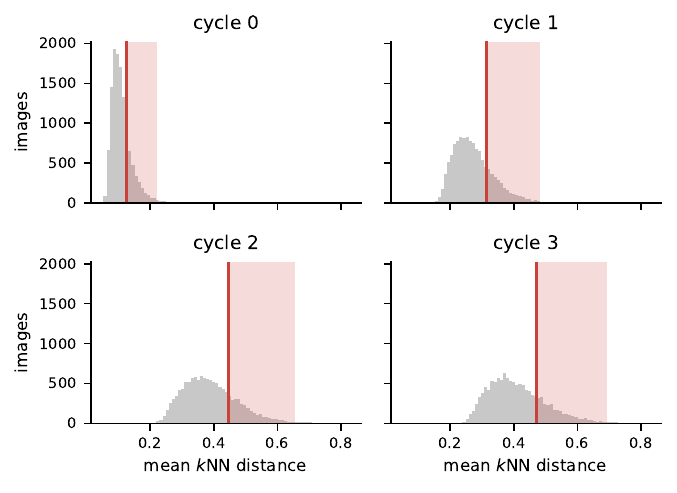}
\caption{Scores of one ImageNet-100-LT run over four repair cycles, 75th to 99th percentile band shaded. The distribution shifts at every cycle, so the band is re-derived each time.}
\label{fig:cycle-histograms-main}
\end{minipage}
\end{figure}
Under the frozen cycle-0 encoder (Figure~\ref{fig:fixed-support}), inserting the generated variants reduces the radius of the treated anchors by 4.7 percent against 2.4 percent for unselected images matched within class on initial radius, a paired difference of $-2.3$ points. Purity falls by 0.6 points at the anchors and by 2.0 at the matched images, so the variants inserted near the anchors mostly share the query's class. The diagnostic uses $k=20$, and per-seed values are in Appendix~\ref{app:geometry}.

\section{Limitations}
\method\ depends on informative encoder neighborhoods in the first cycle and on a frozen generator that imports the biases of its own pretraining data. Repair fidelity, the $\rho$ of Section~\ref{sec:theory}, is not measured in pixel or semantic space, and the deficit of top-tail selection is consistent with low fidelity there. Measured fidelity could set the band edges in place of fixed percentiles. Later cycles select generated images as anchors, 7 to 13 percent at cycle 1 and 34 to 45 percent at cycle 4, with no filter applied. The band, neighborhood size and budget were held at their defaults across all five sources. The loop adds 30 to 45 percent wall-clock time (Appendix~\ref{app:compute}).

\section{Conclusion}
In this paper we introduce \method, which lets a self-supervised learner fill the sparse regions of its own representation and improves transfer from every long-tailed source we tested. Web-scale pretraining corpora are long-tailed in the same way as our sources, and as generators improve, the representation itself could decide at each stage of large-scale pretraining which regions to fill, making data curation part of training. The same loop applies to any modality with a pretrained generator, such as audio, video or molecules, which should be explored in further research.

\clearpage
\section*{AI use statement}
In this work we used generative AI tools for writing and debugging the data generators, the training
and evaluation scripts and the analysis code, for tabulating results, and
for editing the text of the manuscript under the direction of
the authors. We also used them to check the bibliography, and every entry
was subsequently verified against its publisher, proceedings or arXiv record by the authors. We have not used
generative AI tools for generating research ideas. The generated images
that enter pretraining come from the frozen image generator described in
Section~\ref{sec:method}, and no assistant-written text or code is part of
that data. All AI-assisted code was reviewed and
tested by the authors, every number in the paper was checked against the
evaluation outputs, and all text was reviewed and revised by the authors.
We take responsibility for the final content of this work, including text,
claims and artifacts produced with the aid of generative AI.

\section*{Reproducibility statement}
Section~\ref{sec:method} and Algorithm~\ref{alg:bridge} specify the loop, Section~\ref{sec:setup} and Appendix~\ref{app:implementation} give every operating parameter, schedule, generator setting and data split, and Appendix~\ref{app:proofs} contains complete proofs with their assumptions stated in Section~\ref{sec:theory}. Appendix~\ref{app:full-results} reports every per-task number behind the main-text averages.

\section*{Ethics statement}
The work uses public datasets under their release terms and a frozen pretrained image generator, releases no new generator, scraped collection or generated corpus, and involves no human subjects. Generative repair inherits the biases and memorization risks of the generator's pretraining data, and a deployment would require a task-specific review of fairness, privacy and content safety, as Appendix~\ref{app:impact} discusses.

\bibliographystyle{plainnat}
\bibliography{references}

\appendix

\section{Comparison with Related Acquisition Methods}
\label{app:prior-art}
Classical oversampling and TADA receive labels or supervised learning signals (Table~\ref{tab:prior-art}). DOPING uses latent rarity, targets anomaly detection and does not close the loop through a changing SSL representation. GenRep and StableRep make synthetic images or latent neighborhoods the source of contrastive views without allocating a repair budget over an observed skewed corpus. DiffAug is the prior SSL method most similar to ours, since it also iterates between encoder and generator, and it jointly learns a generator for generic positive views without asking where a frozen generator should intervene. AIDE and industrial data engines retrain a model on data collected where it fails a supervised task, using VLM and LLM reasoning, annotation and domain-specific verification. \method\ studies the label-free case in which local support in the current SSL representation is both the diagnostic and the acquisition signal.

\begin{table*}[h]
\centering
\small
\setlength{\tabcolsep}{4pt}
\caption{Closest methods and the dimensions relevant to the acquisition problem. Feedback means that the learner or task model changes the next acquisition decision, and the methods are cited in Section~\ref{sec:background}.}
\label{tab:prior-art}
\begin{adjustbox}{max width=\textwidth}
\begin{tabular}{lllll}
\toprule
Method & Labels or task loss & Acquisition signal & Data action & Feedback \\
\midrule
SMOTE & class labels & class minority & interpolate examples & no \\
DOPING & no labels & latent edge density & latent oversampling & no \\
Concept pruning & no task labels & cluster complexity & remove examples & no \\
DA-Fusion & labels or few-shot set & class and exemplar & image-conditioned diffusion & no \\
TADA & supervised loss dynamics & slow-learned examples & faithful diffusion variants & no \\
GenRep and StableRep & prompts or generator latent & global sampling & synthetic SSL views & no \\
DiffAug & no labels & learned positive-view objective & jointly learned diffusion views & yes \\
AIDE & task labels and failures & VLM and LLM failure analysis & curate, generate, annotate & yes \\
\method & no labels & local support in current SSL space & budgeted local variants & yes \\
\bottomrule
\end{tabular}
\end{adjustbox}
\end{table*}

\section{Representation Diagnostics}
\label{app:geometry}
\begin{table}[t]
\centering
\small
\setlength{\tabcolsep}{5pt}
\caption{Fixed-encoder intervention test behind Figure~\ref{fig:fixed-support}, cycle-0 encoder, $k=20$. Generated images shrink the radius of their own anchors about twice as much as that of matched unselected images, and lower anchor purity less. The last two columns are anchor-minus-matched differences.}
\label{tab:fixed-support}
\begin{tabular}{lrrrrrr}
\toprule
 & \multicolumn{2}{c}{Radius change (\%)} & \multicolumn{2}{c}{Purity change (pts)} & \multicolumn{2}{c}{Difference} \\
Seed & Anchors & Matched & Anchors & Matched & Radius & Purity \\
\midrule
0 & $-5.38$ & $-2.27$ & $-1.57$ & $-2.79$ & $-3.11$ & $+1.22$ \\
1 & $-5.01$ & $-2.35$ & $-0.20$ & $-1.91$ & $-2.66$ & $+1.70$ \\
2 & $-3.62$ & $-2.58$ & $-0.07$ & $-1.17$ & $-1.05$ & $+1.09$ \\
\midrule
Mean & $-4.67$ & $-2.40$ & $-0.61$ & $-1.96$ & $-2.27$ & $+1.34$ \\
\bottomrule
\end{tabular}
\end{table}

\begin{figure}[t]
\centering
\includegraphics[width=0.9\linewidth]{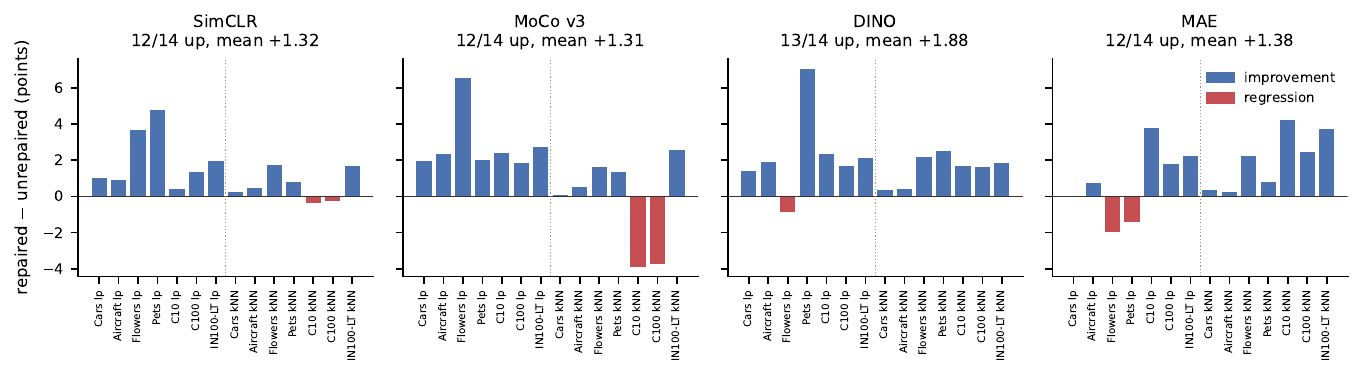}
\caption{Repair against no repair for four objectives on ViT-S, one bar per downstream measurement (linear probe and $k$NN on each of seven datasets), from the paired runs of Table~\ref{tab:main_generalization}. Blue bars are improvements. Red bars are regressions, two for SimCLR and MoCo v3 (CIFAR-10 and CIFAR-100 under $k$NN), one for DINO (Flowers) and two for MAE (Flowers and Pets under the linear probe).}
\label{fig:objective-deltas}
\end{figure}
The fixed-encoder intervention test freezes the cycle-0 encoder and queries the same original images before and after inserting all generated samples, so any change reflects the inserted points alone. First-cycle anchors are matched one-to-one within class to original images with the closest initial $k$NN radius that are never selected in any cycle, and Table~\ref{tab:fixed-support} reports the per-seed values behind Figure~\ref{fig:fixed-support}. Figure~\ref{fig:objective-deltas} draws every paired ViT-S measurement of Table~\ref{tab:main_generalization} as a difference. The learned-space comparison retrains the encoder after the intervention and evaluates the same original-image panel. On that panel MoCo v3 on ViT-S gains $21.97\pm1.58$ in effective rank, $0.296\pm0.019$ in spectral entropy and $3.80\pm0.80$ points of 1-NN accuracy after repair, and the first two survive Holm correction, whereas SimCLR improves transfer while its effective rank falls and its local effect is mixed. We therefore use the fixed-encoder result to locate where the variants land in representation space and the learned-space statistics as objective-dependent diagnostics.

\paragraph{Where the temperature-schedule hybrid helps.} Combining \method\ with TS lowers the seven-task average in every source regime, and it wins 7 of the 70 individual cells (Appendix Tables~\ref{tab:app_sota_linear_full} and~\ref{tab:app_sota_knn_full}). Six of those seven fall on CIFAR-10 and ImageNet-100-LT, the two long-tailed downstream targets, and the largest are $+3.3$ and $+2.8$ points on ImageNet-100-LT and CIFAR-10 under a PASS and a CIFAR-10-LT source. The fine-grained targets lose in almost every case. TS reweights the contrastive temperature toward rare samples, which sharpens the head-to-tail boundary that the long-tailed targets score, and the hybrid gains on those targets. \method\ alone gains more on the four fine-grained targets.

\section{Proofs for the Support-Allocation Framework}
\label{app:proofs}

\begin{proposition}[Closed-form interior optimum]
\label{prop:closed-form}
Let $b(d)=b_0d^{\beta}$ with $b_0>0$ and $\beta\in(0,1]$, and let $\rho(d)=\rho_0e^{-\mu d}$ with $\mu>0$. Then $g$ is strictly decreasing on $(0,\infty)$ for every $\lambda\geq0$, so Proposition~\ref{prop:interior} applies whenever $g$ changes sign on the interval. If $\lambda=0$ the maximizer is $d^{\star}=\beta/\mu$, independent of $b_0$ and $\rho_0$.
\end{proposition}

\begin{proof}[Proof of Proposition~\ref{prop:local-support}]
Sort the distances from $z$ to points in $Z$ and denote their $k$th order statistic by $r_k(z;Z)$. Adding $A$ preserves every old candidate distance and may introduce smaller ones. The $k$th order statistic can therefore only decrease. It decreases strictly when an added distance enters the first $k$ positions and displaces a strictly larger distance.
\end{proof}

\begin{proof}[Proof of Proposition~\ref{prop:sparse-value}]
Let $x=n_j+m_j+\kappa$. Since $f(x)=x^{-\gamma}$ is positive, strictly decreasing and strictly convex for $x>0$, the forward difference $f(x)-f(x+1)$ is positive and strictly decreases with $x$. Multiplication by $p_ja_j>0$ preserves both properties. Equal $p_ja_j$ therefore leaves effective support as the sole determinant of the marginal ordering.
\end{proof}

\begin{proof}[Proof of Proposition~\ref{prop:interior}]
Rewrite the utility as $u(d)=\rho(d)(b(d)+\lambda)-\lambda$. Its derivative is
\[
u'(d)=\rho(d)(b(d)+\lambda)\left[\frac{b'(d)}{b(d)+\lambda}+\frac{\rho'(d)}{\rho(d)}\right]=\rho(d)(b(d)+\lambda)\,g(d).
\]
The factor preceding $g(d)$ is positive. Since $g$ is strictly decreasing and changes sign across the interval, it has exactly one root $d^\star$ by continuity. Hence $u$ increases before $d^\star$ and decreases after it, making $d^\star$ the unique global maximizer.
\end{proof}

\begin{proof}[Proof of Proposition~\ref{prop:closed-form}]
With $b(d)=b_0d^{\beta}$ and $\rho(d)=\rho_0e^{-\mu d}$ we have $\rho'(d)/\rho(d)=-\mu$, so
\[
g(d)=\frac{b_0\beta d^{\beta-1}}{b_0d^{\beta}+\lambda}-\mu=\frac{\beta d^{\beta-1}}{d^{\beta}+c}-\mu,\qquad c=\lambda/b_0\geq0.
\]
Write $h(d)=\beta d^{\beta-1}(d^{\beta}+c)^{-1}$. Differentiating,
\[
h'(d)=\frac{\beta d^{\beta-2}\bigl[(\beta-1)(d^{\beta}+c)-\beta d^{\beta}\bigr]}{(d^{\beta}+c)^{2}}=\frac{\beta d^{\beta-2}\bigl[(\beta-1)c-d^{\beta}\bigr]}{(d^{\beta}+c)^{2}}.
\]
For $\beta\in(0,1]$ we have $(\beta-1)c\leq0<d^{\beta}$ for all $d>0$, so the bracket is strictly negative and $h'(d)<0$. Hence $g=h-\mu$ is strictly decreasing for every $\lambda\geq0$. When $\lambda=0$ we have $c=0$ and $g(d)=\beta/d-\mu$, with unique root is $d^{\star}=\beta/\mu$. Since $g$ is positive below it and negative above it, $d^{\star}$ is the maximizer by the argument of Proposition~\ref{prop:interior}.
\end{proof}

\begin{proof}[Proof of Proposition~\ref{prop:adaptive}]
For region $j$ define the marginal sequence $\delta_{j,t}=U_j(n_j+t)-U_j(n_j+t-1)$ for $t\geq1$, which discrete concavity makes nonincreasing, so every feasible allocation of $R$ repairs selects a prefix of length $m_j$ from each sequence and its utility is the baseline plus the selected prefix marginals. After $t$ greedy steps the selected set is the $t$ largest prefix-feasible marginals, and the next greedy pick is the largest remaining prefix-feasible marginal, so the set stays the top $t+1$ and the claim follows by induction on $t$. No feasible allocation can therefore exceed the greedy total. For the second sentence take two regions with marginal sequences $(3,1,1)$ and $(2,2,2)$ and a budget of three, where the fixed allocation $(3,0)$ built from the initial ordering yields $5$ while greedy yields $3+2+2=7$.
\end{proof}

\begin{proof}[Proof of Proposition~\ref{prop:one-stage}]
Write $f(x)=x^{-\gamma}$. For $x>0$ the mean value theorem gives $f(x)-f(x+1)=\gamma\xi^{-\gamma-1}$ for some $\xi\in(x,x+1)$, so
\[
0<\Delta_j(m_j)=p_ja_j\bigl[f(n_j+m_j+\kappa)-f(n_j+m_j+1+\kappa)\bigr]\leq p_ja_j\gamma(n_j+\kappa)^{-\gamma-1},
\]
using $\xi>n_j+m_j+\kappa\geq n_j+\kappa$ and $m_j\geq0$. Both $\mathcal B(\mathbf m)$ and $\mathcal B(\mathbf m')$ are obtained from the common baseline $\mathcal B(\mathbf 0)$ by subtracting sums of at most $R$ such marginals, since $\sum_jm_j=\sum_jm_j'=R$. Each sum lies in $[0,\,R\gamma\max_jp_ja_j(n_j+\kappa)^{-\gamma-1}]$, and the difference of two numbers drawn from an interval is bounded by its length. Monotonicity of $(n+\kappa)^{-\gamma-1}$ in $n$ gives the $n_{\min}$ form.
\end{proof}

\section{Implementation and Reproducibility}
\label{app:implementation}

\paragraph{Training schedule.} The encoder is warm-started from the previous cycle and is never retrained from scratch. Each cycle uses a fresh optimizer and learning-rate schedule. All earlier real and generated images remain in the source set, so the setting differs from sequential-task learning in which previous data disappear. The cycle ablation holds the total optimizer update budget and the total number of generated images fixed and compares fewer repairs with more learning between them against more frequent repairs with less.

\paragraph{Core configuration.}
The main experiments use ResNet-50 encoders, five cycles of 100 epochs, mean $k$NN distance with $k=100$, 500 anchors per cycle and five variants per anchor. The optimizer is AdamW with weight decay $10^{-4}$, and method-specific learning rates and temperatures follow the settings of each baseline. The paired ViT-S experiments use the reference recipe of each objective. DINO uses two global and eight local crops at $96\times96$ with a bottleneck projection head, and MAE runs five cycles of 80 epochs.

\paragraph{Scoring and selection.}
Scoring uses the mean squared normalized $\ell_2$ distance of Equation~\ref{eq:score}. For unit vectors the squared distance is an affine function of the cosine, so cosine scoring selects the identical anchor set. The mode-window selector restricts the score distribution to the 75th to 99th percentile range, builds a histogram with the automatic bin heuristic, forms a mode-centered candidate pool of size $4B$ and applies greedy farthest-point sampling on normalized features to select the final $B$ anchors.

\paragraph{Generators.}
The main method uses Stable Diffusion 3 Medium in image-to-image mode, where an empty prompt makes the guidance scale inert. The generation ablation replaces it with FLUX or with the oracle, which adds the withheld real images and runs no generator. In the PASS and DiffusionDB experiments the SD3 batch size is 12 to fit the 24-hour wall-time limit. The DiffusionDB source is itself synthetic and shows the stylistic bias of the generator that produced it, so the added variants there are synthetic images derived from synthetic anchors.

\paragraph{Source data.}
PASS and DiffusionDB use \texttt{train[:10000]} with pretraining split override \texttt{[1.0, 0.0, 0.0]}. The long-tailed ImageNet-100, CIFAR-10 and CIFAR-100 sources use their standard long-tailed configurations without the 10{,}000-image restriction. The ImageNet-100 long tail is built from the \texttt{clane9/imagenet-100} train and validation split. Class ranks $1$ to $100$ receive Pareto weights $r^{-1/\alpha}$ with $\alpha=2.5$, the weights are min-max normalized and mapped linearly onto retention counts from 1{,}280 down to 5 images, the resulting per-class ratios are permuted uniformly at random over the classes, and each image is retained by an independent Bernoulli draw with the ratio of its class. The balanced source of Table~\ref{tab:main_controlled_imbalance} reads the same split and replaces the per-class ratios by their mean, so it removes the same expected fraction of images uniformly over classes.

\paragraph{Seeds and reporting.}
All claims rest on three-seed averages reported as mean and standard deviation. Paired comparisons share seeds and, where stated, source checkpoints. Multiple comparisons within one table are corrected by the Holm procedure. No two conditions within a table share an encoder fingerprint.

\section{Compute Resources and Runtime}
\label{app:compute}
Most experiments use one node with two NVIDIA A100 80GB GPUs under a 24-hour wall-time limit. The DINO runs fit on a single 48GB card at the batch size used. The 454 completed runs retained for this paper total approximately 5{,}471 wall-clock hours, of which the baselines and ablations that appear only in the appendix account for 2{,}987. Runs on two-GPU nodes consume roughly twice that in GPU hours, while the paired ViT-S runs hold a single card each. Failed runs, timeouts, cache migration and pilot jobs are excluded from these totals.

Mean wall-clock times over three seeds are 16.9 hours for SimCLR and 21.8 for \method\ on ImageNet-100-LT, 10.2 and 14.9 on PASS, and 11.6 and 16.4 on DiffusionDB. The loop therefore raises wall-clock time by roughly 30 to 45 percent over the corresponding no-repair run, of which the embedding pass is a small fraction and diffusion sampling the remainder. Four repairs add 10{,}000 images, which is 10 percent of a 100{,}000-image source and 1 percent of a one-million-image source. Whether a fixed budget still helps at 100{,}000 or one million images was not tested.

\section{Assets and Terms of Use}
\label{app:assets}
ImageNet-100-LT is derived from ImageNet. The ImageNet access terms restrict the images to non-commercial research and educational use. PASS is released under Creative Commons Attribution 4.0 and DiffusionDB under CC0-1.0. CIFAR-10, CIFAR-100, Stanford Cars, FGVC Aircraft, Oxford Flowers and Oxford-IIIT Pets are used from their official academic releases under their original terms. Stable Diffusion 3 Medium is used under the Stability AI Community License. FLUX.1-dev is used only in the ablation under the FLUX.1-dev Non-Commercial License. We do not redistribute model weights, third-party datasets or generated image corpora.

\section{Broader Impact}
\label{app:impact}
A method that improves representation quality under biased source data can reduce the need for manual relabeling and can make SSL more useful where balanced data are difficult to collect. The method also relies on large image generators and web-sourced datasets, which raise risks of license violations, privacy and unintended memorization or harmful content. Our experiments stay within existing dataset and model release terms, release no new generators or scraped collections, and any deployment would need a review of these risks for its own task.

\section{Full Results}
\label{app:full-results}
\begin{table}[H]
\centering
\tiny
\setlength{\tabcolsep}{3.5pt}
\caption{Full linear-probe transfer results across all source datasets and comparison methods.}
\label{tab:app_sota_linear_full}
\begin{adjustbox}{width=\textwidth}
\begin{tabular}{l l c c c c c c c c c}
\toprule
Source & Method & C10 & C100 & Cars & Aircraft & Flowers & Pets & IN100-LT & Avg. \\
\midrule
ImageNet-100-LT & SimCLR & 73.9 {\scriptsize$\pm$ 0.5} & 46.8 {\scriptsize$\pm$ 0.4} & 13.7 {\scriptsize$\pm$ 0.3} & 19.6 {\scriptsize$\pm$ 0.2} & 35.9 {\scriptsize$\pm$ 7.2} & 38.6 {\scriptsize$\pm$ 1.5} & 50.4 {\scriptsize$\pm$ 0.5} & 39.8 {\scriptsize$\pm$ 1.5} \\
 & TS & 72.5 {\scriptsize$\pm$ 0.4} & 44.0 {\scriptsize$\pm$ 0.5} & 12.5 {\scriptsize$\pm$ 1.2} & 15.0 {\scriptsize$\pm$ 0.1} & 35.7 {\scriptsize$\pm$ 1.8} & 40.6 {\scriptsize$\pm$ 2.2} & \textbf{56.7 {\scriptsize$\pm$ 1.2}} & 39.6 {\scriptsize$\pm$ 1.1} \\
 & SDCLR & 41.1 {\scriptsize$\pm$ 1.7} & 18.0 {\scriptsize$\pm$ 0.6} & 1.4 {\scriptsize$\pm$ 0.3} & 2.3 {\scriptsize$\pm$ 0.5} & 10.8 {\scriptsize$\pm$ 1.7} & 10.4 {\scriptsize$\pm$ 0.3} & 21.9 {\scriptsize$\pm$ 0.6} & 15.1 {\scriptsize$\pm$ 0.8} \\
 & BRIDGE (ours) & \textbf{75.8 {\scriptsize$\pm$ 0.2}} & \textbf{49.3 {\scriptsize$\pm$ 0.2}} & \textbf{16.4 {\scriptsize$\pm$ 1.2}} & \textbf{22.4 {\scriptsize$\pm$ 0.4}} & \textbf{36.8 {\scriptsize$\pm$ 1.9}} & \textbf{42.9 {\scriptsize$\pm$ 1.0}} & 53.8 {\scriptsize$\pm$ 0.1} & \textbf{42.5 {\scriptsize$\pm$ 0.7}} \\
 & BRIDGE+TS (ours) & 74.2 {\scriptsize$\pm$ 0.5} & 45.6 {\scriptsize$\pm$ 0.8} & 9.4 {\scriptsize$\pm$ 0.5} & 15.0 {\scriptsize$\pm$ 0.2} & 25.5 {\scriptsize$\pm$ 2.5} & 33.1 {\scriptsize$\pm$ 2.8} & 48.3 {\scriptsize$\pm$ 0.4} & 35.9 {\scriptsize$\pm$ 1.1} \\
 & BRIDGE+SDCLR (ours) & 43.8 {\scriptsize$\pm$ 1.7} & 18.4 {\scriptsize$\pm$ 0.7} & 2.0 {\scriptsize$\pm$ 0.3} & 2.5 {\scriptsize$\pm$ 0.8} & 10.0 {\scriptsize$\pm$ 0.8} & 12.2 {\scriptsize$\pm$ 3.1} & 21.8 {\scriptsize$\pm$ 0.5} & 15.8 {\scriptsize$\pm$ 1.1} \\
\midrule
CIFAR-10-LT & SimCLR & 67.8 {\scriptsize$\pm$ 0.2} & 38.0 {\scriptsize$\pm$ 1.0} & 6.5 {\scriptsize$\pm$ 0.4} & 16.0 {\scriptsize$\pm$ 0.9} & \textbf{19.2 {\scriptsize$\pm$ 1.6}} & 24.0 {\scriptsize$\pm$ 1.3} & 32.0 {\scriptsize$\pm$ 0.7} & 29.1 {\scriptsize$\pm$ 0.9} \\
 & TS & 68.5 {\scriptsize$\pm$ 1.3} & 36.0 {\scriptsize$\pm$ 1.7} & 4.3 {\scriptsize$\pm$ 0.2} & 10.3 {\scriptsize$\pm$ 0.1} & 13.1 {\scriptsize$\pm$ 2.0} & 19.7 {\scriptsize$\pm$ 2.3} & 31.5 {\scriptsize$\pm$ 1.2} & 26.2 {\scriptsize$\pm$ 1.3} \\
 & SDCLR & 39.6 {\scriptsize$\pm$ 1.1} & 15.6 {\scriptsize$\pm$ 0.2} & 1.1 {\scriptsize$\pm$ 0.1} & 2.4 {\scriptsize$\pm$ 0.9} & 12.0 {\scriptsize$\pm$ 1.6} & 9.1 {\scriptsize$\pm$ 1.8} & 16.5 {\scriptsize$\pm$ 1.6} & 13.8 {\scriptsize$\pm$ 1.1} \\
 & BRIDGE (ours) & 78.1 {\scriptsize$\pm$ 0.3} & \textbf{49.8 {\scriptsize$\pm$ 0.4}} & \textbf{9.2 {\scriptsize$\pm$ 0.8}} & \textbf{17.7 {\scriptsize$\pm$ 1.0}} & 18.2 {\scriptsize$\pm$ 1.1} & \textbf{27.4 {\scriptsize$\pm$ 2.7}} & 39.0 {\scriptsize$\pm$ 0.6} & \textbf{34.2 {\scriptsize$\pm$ 1.0}} \\
 & BRIDGE+TS (ours) & \textbf{78.6 {\scriptsize$\pm$ 0.3}} & 46.1 {\scriptsize$\pm$ 0.3} & 7.5 {\scriptsize$\pm$ 0.9} & 11.1 {\scriptsize$\pm$ 0.7} & 14.5 {\scriptsize$\pm$ 0.6} & 25.1 {\scriptsize$\pm$ 2.7} & \textbf{39.6 {\scriptsize$\pm$ 0.8}} & 31.8 {\scriptsize$\pm$ 0.9} \\
 & BRIDGE+SDCLR (ours) & 40.9 {\scriptsize$\pm$ 0.7} & 17.9 {\scriptsize$\pm$ 0.5} & 1.4 {\scriptsize$\pm$ 0.1} & 2.2 {\scriptsize$\pm$ 0.6} & 11.0 {\scriptsize$\pm$ 2.4} & 10.0 {\scriptsize$\pm$ 2.7} & 21.0 {\scriptsize$\pm$ 0.6} & 14.9 {\scriptsize$\pm$ 1.1} \\
\midrule
CIFAR-100-LT & SimCLR & 59.0 {\scriptsize$\pm$ 1.5} & 31.1 {\scriptsize$\pm$ 1.6} & 4.1 {\scriptsize$\pm$ 0.3} & 11.2 {\scriptsize$\pm$ 1.0} & 12.6 {\scriptsize$\pm$ 3.4} & 20.1 {\scriptsize$\pm$ 1.0} & 28.7 {\scriptsize$\pm$ 1.3} & 23.8 {\scriptsize$\pm$ 1.4} \\
 & TS & 56.3 {\scriptsize$\pm$ 0.2} & 28.0 {\scriptsize$\pm$ 0.1} & 2.8 {\scriptsize$\pm$ 0.6} & 8.5 {\scriptsize$\pm$ 0.2} & 15.5 {\scriptsize$\pm$ 2.5} & 16.2 {\scriptsize$\pm$ 3.1} & 26.2 {\scriptsize$\pm$ 1.8} & 21.9 {\scriptsize$\pm$ 1.2} \\
 & SDCLR & 37.1 {\scriptsize$\pm$ 0.8} & 15.7 {\scriptsize$\pm$ 0.2} & 1.9 {\scriptsize$\pm$ 0.1} & 2.4 {\scriptsize$\pm$ 0.7} & 9.4 {\scriptsize$\pm$ 1.0} & 9.9 {\scriptsize$\pm$ 1.7} & 16.2 {\scriptsize$\pm$ 0.3} & 13.2 {\scriptsize$\pm$ 0.7} \\
 & BRIDGE (ours) & \textbf{74.2 {\scriptsize$\pm$ 0.5}} & \textbf{48.1 {\scriptsize$\pm$ 0.4}} & \textbf{8.3 {\scriptsize$\pm$ 1.0}} & \textbf{18.2 {\scriptsize$\pm$ 0.7}} & \textbf{24.4 {\scriptsize$\pm$ 2.3}} & 27.6 {\scriptsize$\pm$ 0.7} & 39.8 {\scriptsize$\pm$ 1.1} & \textbf{34.4 {\scriptsize$\pm$ 1.0}} \\
 & BRIDGE+TS (ours) & 73.7 {\scriptsize$\pm$ 0.1} & 47.3 {\scriptsize$\pm$ 0.5} & 7.9 {\scriptsize$\pm$ 0.6} & 14.1 {\scriptsize$\pm$ 0.5} & 18.2 {\scriptsize$\pm$ 3.2} & \textbf{28.4 {\scriptsize$\pm$ 0.4}} & \textbf{41.3 {\scriptsize$\pm$ 0.3}} & 33.0 {\scriptsize$\pm$ 0.8} \\
 & BRIDGE+SDCLR (ours) & 42.0 {\scriptsize$\pm$ 1.5} & 16.9 {\scriptsize$\pm$ 0.9} & 1.3 {\scriptsize$\pm$ 0.1} & 2.2 {\scriptsize$\pm$ 0.6} & 10.6 {\scriptsize$\pm$ 0.8} & 8.3 {\scriptsize$\pm$ 0.8} & 20.2 {\scriptsize$\pm$ 0.9} & 14.5 {\scriptsize$\pm$ 0.8} \\
\midrule
PASS-10k & SimCLR & 72.3 {\scriptsize$\pm$ 0.2} & 45.5 {\scriptsize$\pm$ 0.5} & 11.0 {\scriptsize$\pm$ 0.2} & 17.5 {\scriptsize$\pm$ 1.4} & 33.3 {\scriptsize$\pm$ 1.3} & 33.3 {\scriptsize$\pm$ 0.9} & 44.7 {\scriptsize$\pm$ 0.9} & 36.8 {\scriptsize$\pm$ 0.8} \\
 & TS & 71.5 {\scriptsize$\pm$ 0.2} & 43.5 {\scriptsize$\pm$ 0.4} & 10.2 {\scriptsize$\pm$ 1.2} & 17.2 {\scriptsize$\pm$ 0.2} & 33.1 {\scriptsize$\pm$ 2.1} & 36.4 {\scriptsize$\pm$ 1.8} & 49.4 {\scriptsize$\pm$ 0.4} & 37.3 {\scriptsize$\pm$ 0.9} \\
 & SDCLR & 42.2 {\scriptsize$\pm$ 1.1} & 17.7 {\scriptsize$\pm$ 0.8} & 1.8 {\scriptsize$\pm$ 0.3} & 2.2 {\scriptsize$\pm$ 0.8} & 10.6 {\scriptsize$\pm$ 1.3} & 8.2 {\scriptsize$\pm$ 1.6} & 20.8 {\scriptsize$\pm$ 0.4} & 14.8 {\scriptsize$\pm$ 0.9} \\
 & BRIDGE (ours) & \textbf{73.9 {\scriptsize$\pm$ 0.3}} & \textbf{47.1 {\scriptsize$\pm$ 0.4}} & \textbf{13.6 {\scriptsize$\pm$ 0.7}} & \textbf{20.5 {\scriptsize$\pm$ 0.7}} & \textbf{37.7 {\scriptsize$\pm$ 4.0}} & \textbf{37.1 {\scriptsize$\pm$ 0.7}} & 49.1 {\scriptsize$\pm$ 0.5} & \textbf{39.9 {\scriptsize$\pm$ 1.0}} \\
 & BRIDGE+TS (ours) & 68.9 {\scriptsize$\pm$ 0.6} & 40.6 {\scriptsize$\pm$ 0.7} & 12.1 {\scriptsize$\pm$ 0.4} & 14.1 {\scriptsize$\pm$ 0.5} & 34.6 {\scriptsize$\pm$ 2.9} & 31.6 {\scriptsize$\pm$ 1.5} & \textbf{52.4 {\scriptsize$\pm$ 0.6}} & 36.4 {\scriptsize$\pm$ 1.0} \\
 & BRIDGE+SDCLR (ours) & 44.0 {\scriptsize$\pm$ 0.9} & 18.8 {\scriptsize$\pm$ 0.9} & 1.4 {\scriptsize$\pm$ 0.3} & 1.8 {\scriptsize$\pm$ 0.9} & 12.6 {\scriptsize$\pm$ 2.6} & 12.4 {\scriptsize$\pm$ 1.3} & 21.2 {\scriptsize$\pm$ 1.1} & 16.0 {\scriptsize$\pm$ 1.1} \\
\midrule
DiffusionDB-10k & SimCLR & 74.4 {\scriptsize$\pm$ 0.5} & 48.6 {\scriptsize$\pm$ 0.2} & 10.9 {\scriptsize$\pm$ 0.7} & 17.6 {\scriptsize$\pm$ 0.8} & 31.2 {\scriptsize$\pm$ 2.4} & 36.1 {\scriptsize$\pm$ 1.3} & 45.3 {\scriptsize$\pm$ 0.5} & 37.7 {\scriptsize$\pm$ 0.9} \\
 & TS & 72.3 {\scriptsize$\pm$ 0.3} & 44.4 {\scriptsize$\pm$ 0.3} & 10.7 {\scriptsize$\pm$ 0.6} & 15.4 {\scriptsize$\pm$ 0.6} & 33.1 {\scriptsize$\pm$ 1.1} & \textbf{39.9 {\scriptsize$\pm$ 2.7}} & 48.8 {\scriptsize$\pm$ 0.2} & 37.8 {\scriptsize$\pm$ 0.8} \\
 & SDCLR & 42.4 {\scriptsize$\pm$ 0.2} & 18.4 {\scriptsize$\pm$ 0.2} & 1.1 {\scriptsize$\pm$ 0.2} & 2.4 {\scriptsize$\pm$ 0.1} & 13.0 {\scriptsize$\pm$ 3.7} & 8.2 {\scriptsize$\pm$ 1.4} & 21.0 {\scriptsize$\pm$ 0.6} & 15.2 {\scriptsize$\pm$ 0.9} \\
 & BRIDGE (ours) & \textbf{75.9 {\scriptsize$\pm$ 0.4}} & \textbf{49.9 {\scriptsize$\pm$ 0.5}} & \textbf{12.8 {\scriptsize$\pm$ 1.0}} & \textbf{20.0 {\scriptsize$\pm$ 0.4}} & \textbf{35.7 {\scriptsize$\pm$ 1.1}} & 37.7 {\scriptsize$\pm$ 0.8} & 49.1 {\scriptsize$\pm$ 0.5} & \textbf{40.1 {\scriptsize$\pm$ 0.7}} \\
 & BRIDGE+TS (ours) & 67.8 {\scriptsize$\pm$ 0.6} & 39.1 {\scriptsize$\pm$ 0.5} & 9.4 {\scriptsize$\pm$ 1.2} & 12.4 {\scriptsize$\pm$ 0.4} & 30.3 {\scriptsize$\pm$ 3.0} & 33.7 {\scriptsize$\pm$ 1.9} & \textbf{50.8 {\scriptsize$\pm$ 0.6}} & 34.8 {\scriptsize$\pm$ 1.2} \\
 & BRIDGE+SDCLR (ours) & 43.8 {\scriptsize$\pm$ 0.9} & 18.5 {\scriptsize$\pm$ 0.3} & 1.3 {\scriptsize$\pm$ 0.4} & 2.5 {\scriptsize$\pm$ 0.0} & 14.3 {\scriptsize$\pm$ 1.6} & 11.4 {\scriptsize$\pm$ 0.6} & 21.9 {\scriptsize$\pm$ 0.6} & 16.2 {\scriptsize$\pm$ 0.6} \\
\bottomrule
\end{tabular}
\end{adjustbox}
\end{table}

\begin{table}[H]
\centering
\tiny
\setlength{\tabcolsep}{3.5pt}
\caption{Full $k$NN transfer results across all source datasets and comparison methods.}
\label{tab:app_sota_knn_full}
\begin{adjustbox}{width=\textwidth}
\begin{tabular}{l l c c c c c c c c c}
\toprule
Source & Method & C10 & C100 & Cars & Aircraft & Flowers & Pets & IN100-LT & Avg. \\
\midrule
ImageNet-100-LT & SimCLR & 68.6 {\scriptsize$\pm$ 0.4} & 37.3 {\scriptsize$\pm$ 0.2} & 6.0 {\scriptsize$\pm$ 0.7} & 11.2 {\scriptsize$\pm$ 0.4} & 32.2 {\scriptsize$\pm$ 0.7} & 23.6 {\scriptsize$\pm$ 0.3} & 43.9 {\scriptsize$\pm$ 0.9} & 31.8 {\scriptsize$\pm$ 0.5} \\
 & TS & 68.4 {\scriptsize$\pm$ 0.3} & 36.5 {\scriptsize$\pm$ 0.1} & 4.7 {\scriptsize$\pm$ 0.2} & 8.3 {\scriptsize$\pm$ 0.5} & 26.6 {\scriptsize$\pm$ 1.0} & 23.6 {\scriptsize$\pm$ 1.7} & \textbf{52.2 {\scriptsize$\pm$ 0.5}} & 31.5 {\scriptsize$\pm$ 0.6} \\
 & SDCLR & 32.5 {\scriptsize$\pm$ 0.4} & 10.9 {\scriptsize$\pm$ 0.2} & 1.0 {\scriptsize$\pm$ 0.3} & 2.1 {\scriptsize$\pm$ 0.1} & 10.7 {\scriptsize$\pm$ 0.7} & 4.7 {\scriptsize$\pm$ 0.2} & 11.2 {\scriptsize$\pm$ 0.2} & 10.4 {\scriptsize$\pm$ 0.3} \\
 & BRIDGE (ours) & \textbf{69.0 {\scriptsize$\pm$ 0.5}} & \textbf{37.8 {\scriptsize$\pm$ 0.3}} & \textbf{7.0 {\scriptsize$\pm$ 0.2}} & \textbf{11.7 {\scriptsize$\pm$ 0.2}} & \textbf{36.5 {\scriptsize$\pm$ 0.4}} & \textbf{25.3 {\scriptsize$\pm$ 1.0}} & 48.1 {\scriptsize$\pm$ 0.6} & \textbf{33.6 {\scriptsize$\pm$ 0.4}} \\
 & BRIDGE+TS (ours) & 65.4 {\scriptsize$\pm$ 0.2} & 33.6 {\scriptsize$\pm$ 0.3} & 3.1 {\scriptsize$\pm$ 0.1} & 7.1 {\scriptsize$\pm$ 0.3} & 18.2 {\scriptsize$\pm$ 0.2} & 15.6 {\scriptsize$\pm$ 0.7} & 40.8 {\scriptsize$\pm$ 0.5} & 26.3 {\scriptsize$\pm$ 0.3} \\
 & BRIDGE+SDCLR (ours) & 36.3 {\scriptsize$\pm$ 0.8} & 13.0 {\scriptsize$\pm$ 0.5} & 1.1 {\scriptsize$\pm$ 0.3} & 2.4 {\scriptsize$\pm$ 0.2} & 9.7 {\scriptsize$\pm$ 0.3} & 5.4 {\scriptsize$\pm$ 0.4} & 13.9 {\scriptsize$\pm$ 1.0} & 11.7 {\scriptsize$\pm$ 0.5} \\
\midrule
CIFAR-10-LT & SimCLR & 66.9 {\scriptsize$\pm$ 0.4} & 33.4 {\scriptsize$\pm$ 0.3} & 3.7 {\scriptsize$\pm$ 0.4} & 9.8 {\scriptsize$\pm$ 0.7} & 16.5 {\scriptsize$\pm$ 0.8} & 13.9 {\scriptsize$\pm$ 0.8} & 24.0 {\scriptsize$\pm$ 0.8} & 24.0 {\scriptsize$\pm$ 0.6} \\
 & TS & 66.3 {\scriptsize$\pm$ 0.6} & 27.8 {\scriptsize$\pm$ 0.4} & 2.4 {\scriptsize$\pm$ 0.6} & 6.4 {\scriptsize$\pm$ 0.2} & 10.5 {\scriptsize$\pm$ 0.7} & 11.1 {\scriptsize$\pm$ 0.4} & 22.0 {\scriptsize$\pm$ 0.9} & 20.9 {\scriptsize$\pm$ 0.5} \\
 & SDCLR & 30.1 {\scriptsize$\pm$ 0.3} & 10.0 {\scriptsize$\pm$ 0.3} & 1.7 {\scriptsize$\pm$ 0.3} & 2.2 {\scriptsize$\pm$ 0.2} & 10.1 {\scriptsize$\pm$ 0.7} & 4.4 {\scriptsize$\pm$ 0.4} & 9.8 {\scriptsize$\pm$ 0.3} & 9.7 {\scriptsize$\pm$ 0.4} \\
 & BRIDGE (ours) & 74.8 {\scriptsize$\pm$ 0.4} & \textbf{41.0 {\scriptsize$\pm$ 0.4}} & \textbf{4.8 {\scriptsize$\pm$ 0.5}} & \textbf{10.0 {\scriptsize$\pm$ 0.2}} & \textbf{24.2 {\scriptsize$\pm$ 0.6}} & \textbf{16.2 {\scriptsize$\pm$ 0.7}} & \textbf{29.7 {\scriptsize$\pm$ 0.3}} & \textbf{28.7 {\scriptsize$\pm$ 0.4}} \\
 & BRIDGE+TS (ours) & \textbf{77.6 {\scriptsize$\pm$ 0.7}} & 35.3 {\scriptsize$\pm$ 0.5} & 3.1 {\scriptsize$\pm$ 0.3} & 6.6 {\scriptsize$\pm$ 0.4} & 16.3 {\scriptsize$\pm$ 0.6} & 12.5 {\scriptsize$\pm$ 1.0} & 26.7 {\scriptsize$\pm$ 0.1} & 25.5 {\scriptsize$\pm$ 0.5} \\
 & BRIDGE+SDCLR (ours) & 34.4 {\scriptsize$\pm$ 1.2} & 11.4 {\scriptsize$\pm$ 0.6} & 1.2 {\scriptsize$\pm$ 0.2} & 2.2 {\scriptsize$\pm$ 0.2} & 8.8 {\scriptsize$\pm$ 1.5} & 5.0 {\scriptsize$\pm$ 0.2} & 11.7 {\scriptsize$\pm$ 1.1} & 10.7 {\scriptsize$\pm$ 0.7} \\
\midrule
CIFAR-100-LT & SimCLR & 59.3 {\scriptsize$\pm$ 0.8} & 30.1 {\scriptsize$\pm$ 0.9} & 3.0 {\scriptsize$\pm$ 0.3} & 7.4 {\scriptsize$\pm$ 0.4} & 13.7 {\scriptsize$\pm$ 0.2} & 10.8 {\scriptsize$\pm$ 0.4} & 21.8 {\scriptsize$\pm$ 1.1} & 20.9 {\scriptsize$\pm$ 0.6} \\
 & TS & 53.1 {\scriptsize$\pm$ 0.6} & 24.6 {\scriptsize$\pm$ 0.4} & 2.4 {\scriptsize$\pm$ 0.4} & 5.5 {\scriptsize$\pm$ 0.3} & 9.3 {\scriptsize$\pm$ 0.4} & 8.3 {\scriptsize$\pm$ 0.2} & 18.7 {\scriptsize$\pm$ 1.0} & 17.4 {\scriptsize$\pm$ 0.5} \\
 & SDCLR & 25.7 {\scriptsize$\pm$ 1.2} & 8.5 {\scriptsize$\pm$ 0.6} & 0.9 {\scriptsize$\pm$ 0.4} & 1.9 {\scriptsize$\pm$ 0.2} & 7.3 {\scriptsize$\pm$ 0.3} & 3.9 {\scriptsize$\pm$ 0.3} & 8.3 {\scriptsize$\pm$ 1.1} & 8.1 {\scriptsize$\pm$ 0.6} \\
 & BRIDGE (ours) & \textbf{69.3 {\scriptsize$\pm$ 0.5}} & \textbf{39.5 {\scriptsize$\pm$ 0.1}} & \textbf{4.8 {\scriptsize$\pm$ 0.6}} & \textbf{8.9 {\scriptsize$\pm$ 0.1}} & \textbf{21.9 {\scriptsize$\pm$ 0.1}} & \textbf{15.5 {\scriptsize$\pm$ 0.1}} & \textbf{29.1 {\scriptsize$\pm$ 0.7}} & \textbf{27.0 {\scriptsize$\pm$ 0.3}} \\
 & BRIDGE+TS (ours) & 64.3 {\scriptsize$\pm$ 0.4} & 36.6 {\scriptsize$\pm$ 0.4} & 2.9 {\scriptsize$\pm$ 0.3} & 6.3 {\scriptsize$\pm$ 0.1} & 16.4 {\scriptsize$\pm$ 0.5} & 12.3 {\scriptsize$\pm$ 0.0} & 28.2 {\scriptsize$\pm$ 0.6} & 23.9 {\scriptsize$\pm$ 0.3} \\
 & BRIDGE+SDCLR (ours) & 31.9 {\scriptsize$\pm$ 0.4} & 10.4 {\scriptsize$\pm$ 0.5} & 0.8 {\scriptsize$\pm$ 0.1} & 2.1 {\scriptsize$\pm$ 0.5} & 7.5 {\scriptsize$\pm$ 1.6} & 4.5 {\scriptsize$\pm$ 0.1} & 9.1 {\scriptsize$\pm$ 0.7} & 9.5 {\scriptsize$\pm$ 0.6} \\
\midrule
PASS-10k & SimCLR & 67.2 {\scriptsize$\pm$ 0.2} & \textbf{37.0 {\scriptsize$\pm$ 0.3}} & 5.4 {\scriptsize$\pm$ 0.3} & 10.3 {\scriptsize$\pm$ 0.5} & 30.4 {\scriptsize$\pm$ 0.0} & 16.7 {\scriptsize$\pm$ 0.5} & 38.3 {\scriptsize$\pm$ 0.3} & 29.3 {\scriptsize$\pm$ 0.3} \\
 & TS & 65.2 {\scriptsize$\pm$ 0.7} & 34.6 {\scriptsize$\pm$ 0.3} & 4.4 {\scriptsize$\pm$ 0.2} & 7.9 {\scriptsize$\pm$ 0.2} & 24.5 {\scriptsize$\pm$ 0.1} & 14.4 {\scriptsize$\pm$ 0.4} & 39.3 {\scriptsize$\pm$ 0.4} & 27.2 {\scriptsize$\pm$ 0.3} \\
 & SDCLR & 32.4 {\scriptsize$\pm$ 1.5} & 11.5 {\scriptsize$\pm$ 0.8} & 1.3 {\scriptsize$\pm$ 0.2} & 2.0 {\scriptsize$\pm$ 0.1} & 8.6 {\scriptsize$\pm$ 0.7} & 5.2 {\scriptsize$\pm$ 0.2} & 11.1 {\scriptsize$\pm$ 0.8} & 10.3 {\scriptsize$\pm$ 0.6} \\
 & BRIDGE (ours) & \textbf{67.5 {\scriptsize$\pm$ 0.3}} & 36.9 {\scriptsize$\pm$ 0.3} & \textbf{5.8 {\scriptsize$\pm$ 0.5}} & \textbf{10.7 {\scriptsize$\pm$ 0.4}} & \textbf{33.7 {\scriptsize$\pm$ 0.2}} & \textbf{17.8 {\scriptsize$\pm$ 0.6}} & \textbf{41.4 {\scriptsize$\pm$ 0.8}} & \textbf{30.6 {\scriptsize$\pm$ 0.4}} \\
 & BRIDGE+TS (ours) & 58.8 {\scriptsize$\pm$ 0.5} & 27.9 {\scriptsize$\pm$ 0.5} & 3.0 {\scriptsize$\pm$ 0.2} & 6.6 {\scriptsize$\pm$ 0.1} & 23.0 {\scriptsize$\pm$ 0.3} & 13.4 {\scriptsize$\pm$ 0.4} & 38.8 {\scriptsize$\pm$ 0.8} & 24.5 {\scriptsize$\pm$ 0.4} \\
 & BRIDGE+SDCLR (ours) & 37.1 {\scriptsize$\pm$ 2.0} & 13.6 {\scriptsize$\pm$ 1.0} & 1.6 {\scriptsize$\pm$ 0.2} & 2.4 {\scriptsize$\pm$ 0.2} & 10.1 {\scriptsize$\pm$ 0.5} & 6.3 {\scriptsize$\pm$ 0.3} & 13.1 {\scriptsize$\pm$ 0.6} & 12.0 {\scriptsize$\pm$ 0.7} \\
\midrule
DiffusionDB-10k & SimCLR & 67.3 {\scriptsize$\pm$ 0.1} & 36.2 {\scriptsize$\pm$ 0.4} & 4.3 {\scriptsize$\pm$ 0.2} & 8.8 {\scriptsize$\pm$ 0.1} & 26.8 {\scriptsize$\pm$ 0.4} & 17.3 {\scriptsize$\pm$ 0.1} & 36.7 {\scriptsize$\pm$ 0.8} & 28.2 {\scriptsize$\pm$ 0.3} \\
 & TS & 64.7 {\scriptsize$\pm$ 0.4} & 33.6 {\scriptsize$\pm$ 0.3} & 3.1 {\scriptsize$\pm$ 0.5} & 7.3 {\scriptsize$\pm$ 0.3} & 23.2 {\scriptsize$\pm$ 0.4} & 17.0 {\scriptsize$\pm$ 0.0} & 37.8 {\scriptsize$\pm$ 0.6} & 26.7 {\scriptsize$\pm$ 0.4} \\
 & SDCLR & 32.6 {\scriptsize$\pm$ 1.2} & 10.0 {\scriptsize$\pm$ 0.3} & 1.3 {\scriptsize$\pm$ 0.1} & 2.2 {\scriptsize$\pm$ 0.2} & 6.1 {\scriptsize$\pm$ 0.7} & 4.8 {\scriptsize$\pm$ 0.4} & 10.6 {\scriptsize$\pm$ 0.5} & 9.7 {\scriptsize$\pm$ 0.5} \\
 & BRIDGE (ours) & \textbf{68.3 {\scriptsize$\pm$ 0.2}} & \textbf{37.4 {\scriptsize$\pm$ 0.4}} & \textbf{5.4 {\scriptsize$\pm$ 0.3}} & \textbf{9.8 {\scriptsize$\pm$ 0.1}} & \textbf{30.4 {\scriptsize$\pm$ 0.2}} & \textbf{18.6 {\scriptsize$\pm$ 0.2}} & \textbf{39.4 {\scriptsize$\pm$ 1.0}} & \textbf{29.9 {\scriptsize$\pm$ 0.3}} \\
 & BRIDGE+TS (ours) & 57.5 {\scriptsize$\pm$ 0.8} & 26.6 {\scriptsize$\pm$ 0.3} & 3.0 {\scriptsize$\pm$ 0.4} & 6.5 {\scriptsize$\pm$ 0.1} & 21.7 {\scriptsize$\pm$ 0.8} & 15.5 {\scriptsize$\pm$ 0.4} & 37.0 {\scriptsize$\pm$ 0.4} & 24.0 {\scriptsize$\pm$ 0.5} \\
 & BRIDGE+SDCLR (ours) & 35.2 {\scriptsize$\pm$ 0.7} & 11.1 {\scriptsize$\pm$ 0.3} & 1.0 {\scriptsize$\pm$ 0.1} & 2.2 {\scriptsize$\pm$ 0.1} & 8.1 {\scriptsize$\pm$ 1.0} & 5.0 {\scriptsize$\pm$ 0.3} & 11.3 {\scriptsize$\pm$ 0.7} & 10.6 {\scriptsize$\pm$ 0.5} \\
\bottomrule
\end{tabular}
\end{adjustbox}
\end{table}

\begin{table}[H]
\centering
\scriptsize
\setlength{\tabcolsep}{3.5pt}
\caption{Augmentation mechanism ablation with full linear-probe results. The oracle row restores the withheld real images.}
\label{tab:app_generation_linear}
\begin{adjustbox}{width=\textwidth}
\begin{tabular}{l c c c c c c c c}
\toprule
Setting & C10 & C100 & Cars & Aircraft & Flowers & Pets & IN100-LT & Avg. \\
\midrule
Stable Diffusion 3 (default) & \textbf{75.0 {\scriptsize$\pm$ 0.3}} & \textbf{48.5 {\scriptsize$\pm$ 0.2}} & \textbf{16.3 {\scriptsize$\pm$ 0.8}} & \textbf{20.7 {\scriptsize$\pm$ 0.4}} & 35.7 {\scriptsize$\pm$ 0.9} & \textbf{41.8 {\scriptsize$\pm$ 0.8}} & \textbf{52.4 {\scriptsize$\pm$ 0.7}} & \textbf{41.5 {\scriptsize$\pm$ 0.6}} \\
FLUX & 70.7 {\scriptsize$\pm$ 0.4} & 42.1 {\scriptsize$\pm$ 1.4} & 10.3 {\scriptsize$\pm$ 1.3} & 17.5 {\scriptsize$\pm$ 0.9} & 29.4 {\scriptsize$\pm$ 4.6} & 31.5 {\scriptsize$\pm$ 2.9} & 43.7 {\scriptsize$\pm$ 1.4} & 35.0 {\scriptsize$\pm$ 1.9} \\
Oracle (withheld real images restored) & 74.8 {\scriptsize$\pm$ 0.4} & 47.8 {\scriptsize$\pm$ 0.7} & 13.1 {\scriptsize$\pm$ 0.1} & 20.5 {\scriptsize$\pm$ 0.5} & \textbf{40.3 {\scriptsize$\pm$ 6.1}} & 38.1 {\scriptsize$\pm$ 1.5} & 50.3 {\scriptsize$\pm$ 0.4} & 40.7 {\scriptsize$\pm$ 1.4} \\
\bottomrule
\end{tabular}
\end{adjustbox}
\end{table}

\begin{table}[H]
\centering
\scriptsize
\setlength{\tabcolsep}{3.5pt}
\caption{Augmentation mechanism ablation with full $k$NN results. The oracle row restores the withheld real images.}
\label{tab:app_generation_knn}
\begin{adjustbox}{width=\textwidth}
\begin{tabular}{l c c c c c c c c}
\toprule
Setting & C10 & C100 & Cars & Aircraft & Flowers & Pets & IN100-LT & Avg. \\
\midrule
Stable Diffusion 3 (default) & 68.6 {\scriptsize$\pm$ 0.1} & 37.2 {\scriptsize$\pm$ 0.2} & 6.4 {\scriptsize$\pm$ 0.6} & \textbf{11.5 {\scriptsize$\pm$ 0.2}} & \textbf{35.2 {\scriptsize$\pm$ 0.3}} & \textbf{24.3 {\scriptsize$\pm$ 0.6}} & \textbf{46.8 {\scriptsize$\pm$ 0.7}} & \textbf{32.9 {\scriptsize$\pm$ 0.4}} \\
FLUX & 65.8 {\scriptsize$\pm$ 1.0} & 35.4 {\scriptsize$\pm$ 0.8} & 5.7 {\scriptsize$\pm$ 0.7} & 9.4 {\scriptsize$\pm$ 0.9} & 27.7 {\scriptsize$\pm$ 1.0} & 18.8 {\scriptsize$\pm$ 1.3} & 38.7 {\scriptsize$\pm$ 1.2} & 28.8 {\scriptsize$\pm$ 1.0} \\
Oracle (withheld real images restored) & \textbf{68.8 {\scriptsize$\pm$ 0.2}} & \textbf{37.7 {\scriptsize$\pm$ 0.3}} & \textbf{6.6 {\scriptsize$\pm$ 0.5}} & 11.4 {\scriptsize$\pm$ 0.5} & 32.7 {\scriptsize$\pm$ 0.1} & 23.6 {\scriptsize$\pm$ 0.6} & 44.4 {\scriptsize$\pm$ 0.4} & 32.2 {\scriptsize$\pm$ 0.4} \\
\bottomrule
\end{tabular}
\end{adjustbox}
\end{table}

\begin{table}[H]
\centering
\scriptsize
\setlength{\tabcolsep}{3.5pt}
\caption{Selection rule ablation with full linear-probe results.}
\label{tab:app_selection_linear}
\begin{adjustbox}{width=\textwidth}
\begin{tabular}{l c c c c c c c c}
\toprule
Setting & C10 & C100 & Cars & Aircraft & Flowers & Pets & IN100-LT & Avg. \\
\midrule
Mode-window q1-99 + diversity & 75.6 {\scriptsize$\pm$ 0.1} & 49.2 {\scriptsize$\pm$ 0.2} & \textbf{16.9 {\scriptsize$\pm$ 0.7}} & 21.9 {\scriptsize$\pm$ 0.4} & 34.6 {\scriptsize$\pm$ 2.9} & 43.5 {\scriptsize$\pm$ 0.7} & 52.8 {\scriptsize$\pm$ 0.1} & 42.1 {\scriptsize$\pm$ 0.7} \\
Mode-window q50-99 + diversity & 75.5 {\scriptsize$\pm$ 0.1} & \textbf{49.3 {\scriptsize$\pm$ 0.4}} & 15.2 {\scriptsize$\pm$ 0.3} & \textbf{22.4 {\scriptsize$\pm$ 0.6}} & 35.5 {\scriptsize$\pm$ 1.1} & \textbf{43.8 {\scriptsize$\pm$ 3.1}} & 53.4 {\scriptsize$\pm$ 0.9} & 42.2 {\scriptsize$\pm$ 0.9} \\
Mode-window q75-99 + diversity (default) & \textbf{76.0 {\scriptsize$\pm$ 0.3}} & 49.1 {\scriptsize$\pm$ 0.2} & 15.2 {\scriptsize$\pm$ 1.2} & 22.4 {\scriptsize$\pm$ 0.4} & \textbf{36.8 {\scriptsize$\pm$ 1.9}} & 42.9 {\scriptsize$\pm$ 1.0} & \textbf{53.8 {\scriptsize$\pm$ 0.1}} & \textbf{42.3 {\scriptsize$\pm$ 0.7}} \\
Top-tail & 70.8 {\scriptsize$\pm$ 0.5} & 43.1 {\scriptsize$\pm$ 1.0} & 10.0 {\scriptsize$\pm$ 1.7} & 15.9 {\scriptsize$\pm$ 1.3} & 27.7 {\scriptsize$\pm$ 1.0} & 30.5 {\scriptsize$\pm$ 1.2} & 42.2 {\scriptsize$\pm$ 1.4} & 34.3 {\scriptsize$\pm$ 1.2} \\
Uniform & 71.5 {\scriptsize$\pm$ 0.2} & 43.8 {\scriptsize$\pm$ 0.7} & 9.3 {\scriptsize$\pm$ 0.9} & 17.5 {\scriptsize$\pm$ 1.2} & 32.9 {\scriptsize$\pm$ 0.4} & 31.7 {\scriptsize$\pm$ 3.9} & 42.0 {\scriptsize$\pm$ 0.9} & 35.5 {\scriptsize$\pm$ 1.2} \\
\bottomrule
\end{tabular}
\end{adjustbox}
\end{table}

\begin{table}[H]
\centering
\scriptsize
\setlength{\tabcolsep}{3.5pt}
\caption{Selection rule ablation with full $k$NN results.}
\label{tab:app_selection_knn}
\begin{adjustbox}{width=\textwidth}
\begin{tabular}{l c c c c c c c c}
\toprule
Setting & C10 & C100 & Cars & Aircraft & Flowers & Pets & IN100-LT & Avg. \\
\midrule
Mode-window q1-99 + diversity & \textbf{69.2 {\scriptsize$\pm$ 0.3}} & 37.7 {\scriptsize$\pm$ 0.2} & 6.6 {\scriptsize$\pm$ 0.2} & 11.4 {\scriptsize$\pm$ 0.2} & 35.9 {\scriptsize$\pm$ 0.2} & 25.0 {\scriptsize$\pm$ 0.5} & 48.3 {\scriptsize$\pm$ 0.3} & 33.4 {\scriptsize$\pm$ 0.3} \\
Mode-window q50-99 + diversity & 68.7 {\scriptsize$\pm$ 0.1} & \textbf{38.0 {\scriptsize$\pm$ 0.2}} & \textbf{7.3 {\scriptsize$\pm$ 0.2}} & 11.3 {\scriptsize$\pm$ 0.2} & \textbf{36.5 {\scriptsize$\pm$ 0.1}} & 24.8 {\scriptsize$\pm$ 0.8} & \textbf{48.5 {\scriptsize$\pm$ 0.4}} & 33.6 {\scriptsize$\pm$ 0.3} \\
Mode-window q75-99 + diversity (default) & 69.0 {\scriptsize$\pm$ 0.5} & 37.8 {\scriptsize$\pm$ 0.3} & 6.7 {\scriptsize$\pm$ 0.3} & \textbf{11.7 {\scriptsize$\pm$ 0.2}} & 36.5 {\scriptsize$\pm$ 0.4} & \textbf{25.3 {\scriptsize$\pm$ 1.0}} & 48.1 {\scriptsize$\pm$ 0.6} & \textbf{33.6 {\scriptsize$\pm$ 0.5}} \\
Top-tail & 65.7 {\scriptsize$\pm$ 0.8} & 34.4 {\scriptsize$\pm$ 0.2} & 4.1 {\scriptsize$\pm$ 0.4} & 9.6 {\scriptsize$\pm$ 0.2} & 23.9 {\scriptsize$\pm$ 0.9} & 18.1 {\scriptsize$\pm$ 0.8} & 35.4 {\scriptsize$\pm$ 1.2} & 27.3 {\scriptsize$\pm$ 0.6} \\
Uniform & 66.1 {\scriptsize$\pm$ 1.0} & 35.7 {\scriptsize$\pm$ 0.3} & 4.9 {\scriptsize$\pm$ 0.1} & 10.2 {\scriptsize$\pm$ 0.7} & 25.3 {\scriptsize$\pm$ 0.4} & 19.0 {\scriptsize$\pm$ 0.8} & 37.4 {\scriptsize$\pm$ 0.6} & 28.4 {\scriptsize$\pm$ 0.6} \\
\bottomrule
\end{tabular}
\end{adjustbox}
\end{table}

\begin{table}[H]
\centering
\scriptsize
\setlength{\tabcolsep}{3.5pt}
\caption{Cycle count ablation with full linear-probe results.}
\label{tab:app_cycles_linear}
\begin{adjustbox}{width=\textwidth}
\begin{tabular}{l c c c c c c c c}
\toprule
Setting & C10 & C100 & Cars & Aircraft & Flowers & Pets & IN100-LT & Avg. \\
\midrule
2 cycles & 68.9 {\scriptsize$\pm$ 0.6} & 41.6 {\scriptsize$\pm$ 1.3} & 10.6 {\scriptsize$\pm$ 1.2} & 16.4 {\scriptsize$\pm$ 0.4} & 30.7 {\scriptsize$\pm$ 4.5} & 33.2 {\scriptsize$\pm$ 0.9} & 43.3 {\scriptsize$\pm$ 0.6} & 35.0 {\scriptsize$\pm$ 1.4} \\
5 cycles (default) & \textbf{76.0 {\scriptsize$\pm$ 0.3}} & \textbf{49.1 {\scriptsize$\pm$ 0.2}} & \textbf{15.2 {\scriptsize$\pm$ 1.2}} & \textbf{22.4 {\scriptsize$\pm$ 0.4}} & \textbf{36.8 {\scriptsize$\pm$ 1.9}} & \textbf{42.9 {\scriptsize$\pm$ 1.0}} & \textbf{53.8 {\scriptsize$\pm$ 0.1}} & \textbf{42.3 {\scriptsize$\pm$ 0.7}} \\
10 cycles & 70.5 {\scriptsize$\pm$ 1.2} & 43.8 {\scriptsize$\pm$ 0.6} & 9.2 {\scriptsize$\pm$ 0.9} & 17.2 {\scriptsize$\pm$ 1.6} & 24.8 {\scriptsize$\pm$ 5.2} & 31.3 {\scriptsize$\pm$ 1.7} & 41.9 {\scriptsize$\pm$ 0.6} & 34.1 {\scriptsize$\pm$ 1.7} \\
\bottomrule
\end{tabular}
\end{adjustbox}
\end{table}

\begin{table}[H]
\centering
\scriptsize
\setlength{\tabcolsep}{3.5pt}
\caption{Cycle count ablation with full $k$NN results.}
\label{tab:app_cycles_knn}
\begin{adjustbox}{width=\textwidth}
\begin{tabular}{l c c c c c c c c}
\toprule
Setting & C10 & C100 & Cars & Aircraft & Flowers & Pets & IN100-LT & Avg. \\
\midrule
2 cycles & 65.3 {\scriptsize$\pm$ 0.4} & 34.2 {\scriptsize$\pm$ 0.5} & 5.0 {\scriptsize$\pm$ 0.7} & 10.0 {\scriptsize$\pm$ 0.2} & 26.1 {\scriptsize$\pm$ 0.8} & 18.7 {\scriptsize$\pm$ 0.6} & 35.9 {\scriptsize$\pm$ 0.3} & 27.9 {\scriptsize$\pm$ 0.5} \\
5 cycles (default) & \textbf{69.0 {\scriptsize$\pm$ 0.5}} & \textbf{37.8 {\scriptsize$\pm$ 0.3}} & \textbf{6.7 {\scriptsize$\pm$ 0.3}} & \textbf{11.7 {\scriptsize$\pm$ 0.2}} & \textbf{36.5 {\scriptsize$\pm$ 0.4}} & \textbf{25.3 {\scriptsize$\pm$ 1.0}} & \textbf{48.1 {\scriptsize$\pm$ 0.6}} & \textbf{33.6 {\scriptsize$\pm$ 0.5}} \\
10 cycles & 66.4 {\scriptsize$\pm$ 0.5} & 35.4 {\scriptsize$\pm$ 0.3} & 4.8 {\scriptsize$\pm$ 0.3} & 10.4 {\scriptsize$\pm$ 0.6} & 25.2 {\scriptsize$\pm$ 0.8} & 18.6 {\scriptsize$\pm$ 0.9} & 36.0 {\scriptsize$\pm$ 0.7} & 28.1 {\scriptsize$\pm$ 0.6} \\
\bottomrule
\end{tabular}
\end{adjustbox}
\end{table}

\begin{table}[H]
\centering
\scriptsize
\setlength{\tabcolsep}{3.5pt}
\caption{Backbone architecture ablation with full linear-probe results.}
\label{tab:app_architecture_linear}
\begin{adjustbox}{width=\textwidth}
\begin{tabular}{l c c c c c c c c}
\toprule
Setting & C10 & C100 & Cars & Aircraft & Flowers & Pets & IN100-LT & Avg. \\
\midrule
ResNet-18 & 68.8 {\scriptsize$\pm$ 0.1} & 41.7 {\scriptsize$\pm$ 0.2} & 13.5 {\scriptsize$\pm$ 0.3} & 19.7 {\scriptsize$\pm$ 0.6} & 38.7 {\scriptsize$\pm$ 3.4} & 38.4 {\scriptsize$\pm$ 2.6} & 50.2 {\scriptsize$\pm$ 0.2} & 38.7 {\scriptsize$\pm$ 1.1} \\
ResNet-50 (default) & \textbf{76.0 {\scriptsize$\pm$ 0.3}} & \textbf{49.1 {\scriptsize$\pm$ 0.2}} & \textbf{15.2 {\scriptsize$\pm$ 1.2}} & \textbf{22.4 {\scriptsize$\pm$ 0.4}} & 36.8 {\scriptsize$\pm$ 1.9} & \textbf{42.9 {\scriptsize$\pm$ 1.0}} & \textbf{53.8 {\scriptsize$\pm$ 0.1}} & \textbf{42.3 {\scriptsize$\pm$ 0.7}} \\
ViT-S & 62.7 {\scriptsize$\pm$ 0.4} & 37.6 {\scriptsize$\pm$ 0.6} & 8.8 {\scriptsize$\pm$ 0.5} & 12.2 {\scriptsize$\pm$ 0.2} & 35.1 {\scriptsize$\pm$ 4.2} & 30.2 {\scriptsize$\pm$ 1.6} & 42.5 {\scriptsize$\pm$ 1.2} & 32.7 {\scriptsize$\pm$ 1.3} \\
ViT-B & 66.0 {\scriptsize$\pm$ 2.3} & 40.6 {\scriptsize$\pm$ 2.3} & 8.3 {\scriptsize$\pm$ 0.5} & 14.3 {\scriptsize$\pm$ 0.8} & \textbf{38.8 {\scriptsize$\pm$ 2.6}} & 31.3 {\scriptsize$\pm$ 2.1} & 44.5 {\scriptsize$\pm$ 1.7} & 34.8 {\scriptsize$\pm$ 1.8} \\
\bottomrule
\end{tabular}
\end{adjustbox}
\end{table}

\begin{table}[H]
\centering
\scriptsize
\setlength{\tabcolsep}{3.5pt}
\caption{Backbone architecture ablation with full $k$NN results.}
\label{tab:app_architecture_knn}
\begin{adjustbox}{width=\textwidth}
\begin{tabular}{l c c c c c c c c}
\toprule
Setting & C10 & C100 & Cars & Aircraft & Flowers & Pets & IN100-LT & Avg. \\
\midrule
ResNet-18 & 66.3 {\scriptsize$\pm$ 0.2} & 35.7 {\scriptsize$\pm$ 0.3} & 5.9 {\scriptsize$\pm$ 0.2} & 11.1 {\scriptsize$\pm$ 0.4} & 33.5 {\scriptsize$\pm$ 0.1} & 23.3 {\scriptsize$\pm$ 0.5} & 44.5 {\scriptsize$\pm$ 0.7} & 31.5 {\scriptsize$\pm$ 0.4} \\
ResNet-50 (default) & 69.0 {\scriptsize$\pm$ 0.5} & 37.8 {\scriptsize$\pm$ 0.3} & \textbf{6.7 {\scriptsize$\pm$ 0.3}} & \textbf{11.7 {\scriptsize$\pm$ 0.2}} & 36.5 {\scriptsize$\pm$ 0.4} & \textbf{25.3 {\scriptsize$\pm$ 1.0}} & \textbf{48.1 {\scriptsize$\pm$ 0.6}} & \textbf{33.6 {\scriptsize$\pm$ 0.5}} \\
ViT-S & 67.3 {\scriptsize$\pm$ 0.4} & 37.5 {\scriptsize$\pm$ 0.3} & 5.3 {\scriptsize$\pm$ 0.2} & 7.6 {\scriptsize$\pm$ 0.1} & 36.2 {\scriptsize$\pm$ 0.7} & 20.1 {\scriptsize$\pm$ 1.0} & 42.1 {\scriptsize$\pm$ 1.0} & 30.9 {\scriptsize$\pm$ 0.5} \\
ViT-B & \textbf{69.3 {\scriptsize$\pm$ 0.8}} & \textbf{40.1 {\scriptsize$\pm$ 0.8}} & 6.0 {\scriptsize$\pm$ 0.2} & 8.3 {\scriptsize$\pm$ 0.2} & \textbf{39.8 {\scriptsize$\pm$ 0.8}} & 21.9 {\scriptsize$\pm$ 1.4} & 45.1 {\scriptsize$\pm$ 0.7} & 32.9 {\scriptsize$\pm$ 0.7} \\
\bottomrule
\end{tabular}
\end{adjustbox}
\end{table}

\begin{table*}[t]
\centering
\small
\setlength{\tabcolsep}{4pt}
\caption{Linear-probe accuracy of the paired ViT-S runs of Table~\ref{tab:main_generalization} on every task, mean $\pm$ standard deviation over three seeds. MAE runs five cycles of 80 epochs, the other objectives five of 100.}
\label{tab:generalization-vits-full-linear}
\begin{adjustbox}{max width=\textwidth}
\begin{tabular}{lccccccc}
\toprule
Method & Cars & Aircraft & Flowers & Pets & CIFAR-10 & CIFAR-100 & IN100-LT \\
\midrule
SimCLR, SSL baseline & $7.15\pm0.19$ & $10.92\pm0.40$ & $32.90\pm2.72$ & $25.00\pm1.25$ & $60.46\pm0.46$ & $34.96\pm0.20$ & $41.95\pm1.43$ \\
SimCLR, with \method & $8.18\pm0.42$ & $11.82\pm0.43$ & $36.60\pm2.85$ & $29.80\pm2.98$ & $60.85\pm0.35$ & $36.33\pm0.14$ & $43.93\pm1.09$ \\
MoCo v3, SSL baseline & $7.57\pm0.23$ & $11.05\pm1.31$ & $30.28\pm1.36$ & $29.26\pm3.13$ & $59.96\pm0.39$ & $34.47\pm0.01$ & $44.79\pm1.17$ \\
MoCo v3, with \method & $9.55\pm0.34$ & $13.42\pm0.98$ & $36.82\pm6.07$ & $31.25\pm4.00$ & $62.36\pm0.46$ & $36.30\pm0.13$ & $47.55\pm0.86$ \\
DINO, SSL baseline & $8.10\pm0.25$ & $11.40\pm1.03$ & $38.13\pm5.24$ & $27.54\pm2.34$ & $58.93\pm0.94$ & $33.56\pm1.75$ & $44.63\pm1.47$ \\
DINO, with \method & $9.49\pm0.88$ & $13.28\pm0.58$ & $37.25\pm2.61$ & $34.60\pm3.62$ & $61.30\pm0.07$ & $35.23\pm0.31$ & $46.78\pm0.77$ \\
MAE, SSL baseline & $1.54\pm0.19$ & $2.44\pm1.00$ & $10.46\pm5.88$ & $10.05\pm3.03$ & $37.69\pm3.73$ & $13.20\pm1.94$ & $19.93\pm2.13$ \\
MAE, with \method & $1.56\pm0.43$ & $3.18\pm0.70$ & $8.50\pm2.26$ & $8.61\pm2.20$ & $41.46\pm2.20$ & $15.00\pm2.40$ & $22.19\pm2.35$ \\
\bottomrule
\end{tabular}
\end{adjustbox}
\end{table*}

\begin{table*}[t]
\centering
\small
\setlength{\tabcolsep}{4pt}
\caption{$k$NN accuracy of the paired ViT-S runs of Table~\ref{tab:main_generalization} on every task, mean $\pm$ standard deviation over three seeds. MAE runs five cycles of 80 epochs, the other objectives five of 100.}
\label{tab:generalization-vits-full-knn}
\begin{adjustbox}{max width=\textwidth}
\begin{tabular}{lccccccc}
\toprule
Method & Cars & Aircraft & Flowers & Pets & CIFAR-10 & CIFAR-100 & IN100-LT \\
\midrule
SimCLR, SSL baseline & $6.26\pm0.17$ & $7.09\pm0.17$ & $36.53\pm0.89$ & $19.16\pm0.64$ & $66.05\pm0.48$ & $36.72\pm0.36$ & $41.21\pm0.72$ \\
SimCLR, with \method & $6.52\pm0.22$ & $7.56\pm0.23$ & $38.29\pm0.83$ & $19.93\pm0.53$ & $65.70\pm0.85$ & $36.46\pm0.33$ & $42.91\pm0.78$ \\
MoCo v3, SSL baseline & $6.65\pm0.38$ & $7.71\pm0.16$ & $38.91\pm1.03$ & $23.56\pm1.12$ & $69.29\pm0.25$ & $39.98\pm0.44$ & $45.14\pm0.46$ \\
MoCo v3, with \method & $6.74\pm0.26$ & $8.21\pm0.18$ & $40.51\pm1.45$ & $24.91\pm0.45$ & $65.41\pm0.21$ & $36.24\pm0.46$ & $47.71\pm1.53$ \\
DINO, SSL baseline & $6.05\pm0.27$ & $7.73\pm0.52$ & $38.52\pm1.19$ & $19.26\pm0.16$ & $62.69\pm2.42$ & $33.98\pm2.80$ & $45.28\pm0.98$ \\
DINO, with \method & $6.39\pm0.22$ & $8.16\pm0.26$ & $40.72\pm0.31$ & $21.79\pm1.15$ & $64.38\pm3.51$ & $35.59\pm2.51$ & $47.14\pm0.29$ \\
MAE, SSL baseline & $1.00\pm0.14$ & $2.24\pm0.29$ & $5.30\pm0.52$ & $4.58\pm0.17$ & $31.34\pm1.65$ & $10.96\pm0.44$ & $11.75\pm0.38$ \\
MAE, with \method & $1.38\pm0.37$ & $2.50\pm0.70$ & $7.54\pm2.13$ & $5.35\pm0.76$ & $35.58\pm7.14$ & $13.44\pm3.46$ & $15.46\pm2.68$ \\
\bottomrule
\end{tabular}
\end{adjustbox}
\end{table*}

\section{Algorithm}
\begin{algorithm}[H]
\caption{\method\ training loop with mode-window selection}
\label{alg:bridge}
\begin{algorithmic}[1]
\STATE Initialize source set $D^{(0)}$, encoder $f_{\theta^{(0)}}$, frozen generator $g_{\phi}$, cycle lengths $\{E_c\}_{c=0}^{C-1}$
\FOR{$c = 0, \dots, C-1$}
    \STATE Continue training $f_{\theta^{(c)}}$ on $D^{(c)}$ for $E_c$ epochs with a fresh optimizer schedule
    \IF{$c<C-1$}
        \STATE Compute normalized embeddings $Z^{(c)} = \{f_{\theta^{(c)}}(x_i) \mid x_i \in D^{(c)}\}$
        \STATE Compute mean $k$NN scores $\{d_i^{(c)}\}$ from $Z^{(c)}$
        \STATE Restrict to the 75th to 99th percentile band and locate the modal bin center $\bar d^{(c)}$
        \STATE Form the candidate pool $P^{(c)}$ of the $\alpha B$ images with scores closest to $\bar d^{(c)}$
        \STATE Select $S^{(c)} \subseteq P^{(c)}$ with $|S^{(c)}| = B$ by greedy farthest-point sampling in embedding space
        \STATE Form $\widetilde{D}^{(c)} = \{g_{\phi}(x_i, \epsilon_j) \mid i \in S^{(c)},\, j \in [G]\}$ and set $D^{(c+1)} = D^{(c)} \cup \widetilde{D}^{(c)}$
    \ENDIF
\ENDFOR
\STATE Return the final encoder $f_{\theta^{(C-1)}}$
\end{algorithmic}
\end{algorithm}

\FloatBarrier
\end{document}